\documentclass[final]{cvpr}

\usepackage{times}
\usepackage{epsfig}
\usepackage{graphicx}
\usepackage{amsmath}
\usepackage{amssymb}

\usepackage{microtype}
\usepackage{subfigure}
\usepackage{booktabs}
\usepackage{tablefootnote}

\usepackage{url}

\usepackage{multirow}
\usepackage{bm}
\usepackage{amsthm}

\usepackage{paralist}
\usepackage{enumitem}
\usepackage{makecell}
\usepackage{paralist}

\usepackage{algorithmic}
\usepackage[ruled]{algorithm}

\newtheorem{theorem}{Theorem}
\newtheorem{lemma}{Lemma}

\newtheorem{assumption}{Assumption}

\usepackage[pagebackref=true,breaklinks=true,colorlinks,bookmarks=false]{hyperref}

\def\confYear{CVPR 2022}

\begin{document}

\title{Bandits for Structure Perturbation-based Black-box Attacks to Graph Neural Networks with Theoretical Guarantees}

\author{
Binghui Wang\footnotemark[1], Youqi Li\footnotemark[2], and Pan Zhou\footnotemark[3]\\
\footnotemark[1]\,\,{\scriptsize Department of Computer Science, Illinois Institute of Technology}\\
\footnotemark[2]\,\,{\scriptsize School of Cyberspace Science and Technology, and School of Computer Science, Beijing Institute of Technology}\\
\footnotemark[3]\,\,{\scriptsize Hubei Engineering Research Center on Big Data Security, School of Cyber Science and Engineering, Huazhong University of Science and Technology}\\
Email: \footnotemark[1]\,\,{\tt\small  bwang70@iit.edu}, \footnotemark[2]\,\,{\tt\small liyouqi@bit.edu.cn}, \footnotemark[3]\,\,{\tt\small panzhou@hust.edu.cn}
\vspace{-5mm}
}

\maketitle
\pagestyle{empty} 
\thispagestyle{empty}

\renewcommand{\thefootnote}{\fnsymbol{footnote}}
\footnotetext[3]{Corresponding author.}
\renewcommand{\thefootnote}{\arabic{footnote}}

\begin{abstract}
Graph neural networks (GNNs) have achieved state-of-the-art performance in many graph-based tasks such as node classification and graph classification. However, many recent works have demonstrated that an attacker can mislead GNN models by slightly perturbing the graph structure. Existing attacks to GNNs are either under the less practical threat model where the attacker is assumed to access the GNN model parameters, or under the practical black-box threat model but consider perturbing node features that are shown to be not enough effective. In this paper, we aim to bridge this gap and consider black-box attacks to GNNs with structure perturbation as well as with theoretical guarantees. We propose to address this challenge through bandit techniques. Specifically, we formulate our attack as an online optimization with bandit feedback. This original problem is essentially NP-hard due to the fact that perturbing the graph structure is a binary optimization problem. We then propose an online attack based on bandit optimization which is proven to be {sublinear} to the query number $T$, i.e., $\mathcal{O}(\sqrt{N}T^{3/4})$ where $N$ is the number of nodes in the graph. Finally, we evaluate our proposed attack by conducting experiments over multiple datasets and GNN models. The experimental results on various citation graphs and image graphs show that our attack is both effective and efficient. Source code is available at~{\url{https://github.com/Metaoblivion/Bandit_GNN_Attack}} 
\end{abstract}

\section{Introduction}
Graph neural networks (GNNs) have been emerging as the most prominent methodology for learning with graphs, such as 
social networks, chemical networks, superpixel graphs, etc. GNNs have also advanced many graph-related applications including but not limited to drug discovery~\cite{shi2020graphaf}, fake news detection on social media~\cite{monti2019fake}, traffic forecasting~\cite{yu2018spatio}, and superpixel graph classification~\cite{dwivedi2020benchmarkgnns}. However, recent works have shown that GNNs are vulnerable to adversarial attacks~\cite{dai2018adversarial,zugner2018adversarial,wu2019adversarial,zugner2019adversarial,wang2019attacking,xu2019topology,sun2020adversarial,ma2020towards}. That is, an attacker can easily fool a GNN model by slightly perturbing the graph structure (e.g., injecting new fake edges into the graph or deleting the existing edges from the graph) or perturbing the node features. Most of the existing attacks to GNNs essentially rely on white-box or gray-box threat model \cite{dai2018adversarial,zugner2018adversarial,wu2019adversarial,zugner2019adversarial,wang2019attacking,xu2019topology,sun2020adversarial}. An attacker can not only obtain the predictions generated by the targeted GNN model, but also know the whole (i.e., in white-box) or partial (i.e., in gray-box) GNNs' inner parameters and network structure. These threat models enable the attacker to derive the true gradients that can be used to construct an (almost) {optimal} edge/feature perturbation via first-order optimization approaches, e.g., projected gradient descent (PDG).

In practice, however, an attacker often has limited knowledge about the GNN model. For instance, many models are deployed as an API due to the commercial value. In these practical scenarios, an attacker can only obtain the model predictions by querying the API, while not knowing the model's other information. An attack based on such a realistic threat model is called a \emph{black-box attack}, which significantly raises the bar for the attacker as he cannot obtain the gradient information. A recent work~\cite{ma2020towards} performs black-box attacks against GNNs. However, this work has two key drawbacks. First, it assumes that the attacker can only perturb the (continuous) node features. Existing works (e.g., \cite{zugner2018adversarial}) have shown that feature perturbation-based attacks to GNNs are significantly less effective than structure perturbation-based attacks. Second, the attack is implemented via a heuristic greedy algorithm, which has no  theoretically guaranteed attack performance. Note that black-box attacks are classified as \emph{soft-label} black-box attacks \cite{ilyas2018black,li2019nattack} and \emph{hard-label} \cite{cheng2018query} black-box attacks. The former means an attacker knows the confidence scores when querying a target model, while the latter means an attacker only knows the predicted label. 

In this paper, we consider soft-label black-box attacks to GNNs with \emph{structure perturbation}. However, such a new attack setting is much more challenging, as finding the optimal structure perturbation is essentially an NP-hard problem (i.e., a binary optimization problem) and the attacker only obtains the predictions via querying the model. We take the first step to solve the structure perturbation-based black-box attacks to GNNs  with theoretical guarantees. Specifically, we first reformulate our attack as a bandit optimization (i.e., online optimization with bandit feedback) problem, which characterizes the attacker's query process on the black-box GNN model and captures the unknown gradients. Then, we handle the binary constraint of the discrete structure perturbation and integrate it into our bandit-based attack objective. Next, we design an efficient and effective online attack to GNNs. Finally, we theoretically analyze our attack. Our key contributions are summarized as follows: 
\begin{itemize}[leftmargin=*,nolistsep,nosep]
\item We design the first {theoretically guaranteed} structure perturbation-based black-box attacks to GNNs. 
\item We prove that our bandit-based attack algorithm theoretically yields a {sublinear} regret bound $\mathcal{O}(\sqrt{N}T^{3/4})$ within $T$ queries for attacking a graph with $N$ nodes. 
\item We conduct extensive experiments to evaluate our attack over  multiple graph datasets and GNN models and demonstrate the effectiveness and efficiency of our attack. 
\end{itemize}

\section{Preliminaries and Problem Formulation}
\subsection{Graph neural networks}
Let $G=(\mathcal{V}, \mathcal{E}, \bm{X})$ be a graph, where $u \in \mathcal{V}$ is a node, $(u, v) \in \mathcal{E}$ is an edge between $u$ and $v$, and $\bm{X} = [\bm{x}_1; \bm{x}_2; \cdots; \bm{x}_{N}] \in \mathbb{R}^{N \times d} $ is the node feature matrix. We denote $\bm{a}_v = [{a}_{v1}; {a}_{v2}; \cdots; {a}_{vN}] \in \{0,1\}^N$ as the adjacency vector of node $v$. $N = |\mathcal{V}|$ and $M = |\mathcal{E}|$ are the number of nodes and edges, respectively. We denote $d_u$ and $\mathcal{N}(u)$ as $u$'s node degree and the neighborhood set of $u$. We consider GNNs for node classification in this paper\footnote{Our attack can be naturally generalized to GNNs for graph classification. We discuss this in Section 3.1.}. In this context, each node $u \in \mathcal{V}$ has a label $y_u$ from a label set $\mathcal{Y} = \{1, 2, \cdots, L_C \}$. Given a set of $\mathcal{V}_L \subset \mathcal{V}$ labeled nodes $\{(\bm{x}_u, y_u)\}_{u \in \mathcal{V}_L}$ as the training set, GNN for node classification is to learn a node classifier that maps each unlabeled node $u \in \mathcal{V} \setminus \mathcal{V}_L$ to a class $y \in \mathcal{Y}$ based on the graph $G$. 

Generally speaking, GNN consists of two main steps: \emph{neighborhood aggregation} and \emph{node representation update}. Suppose a GNN has $K$ layers. We denote $v$'s representation  in the $k$-th layer as $\bm{h}_v^{(k)}$, with $\bm{h}_v^{(0)} = \bm{x}_v$. In the neighborhood aggregation, GNN obtains the representation $\bm{l}_v^{(k)}$ by aggregating representations of $v$'s neighbors in the $(k-1)$-th layer as $\bm{l}_v^{(k)} =  \textrm{AGG} \big( \big\{ \bm{h}_u^{(k-1)}: u \in \mathcal{N}(v) \big\} \big)$. In the node representation update, GNN updates $v$'s representation at the $k$-th layer via 
combining $v$'s previous layer's representation $\bm{h}_v^{(k-1)}$ with the aggregated neighborhood's representations $\bm{l}_v^{(k)}$: $\bm{h}_v^{(k)} = \textrm{UPDATE}\big(\bm{h}_v^{(k-1)}, \bm{l}_v^{(k)} \big)$.

Different GNNs use different $\textrm{AGG}$ and $\textrm{UPDATE}$ functions. For instance, in Graph Convolutional Network (GCN)~\cite{kipf2017semi}, $\textrm{AGG}$ is the element-wise mean pooling function and $\textrm{UPDATE}$ is the $\textrm{ReLU}$ activation function. More specifically, it has the following form: $\bm{h}_v^{(k)} = \textrm{ReLU}\Big( \bm{W}^{(k)}\big( \sum_{u \in \mathcal{N}(v)} d_u^{-1/2} d_v^{-1/2} \bm{h}_{u}^{(k-1)} \big) \Big)$, where $\bm{W}^{(k)}$ is the parameters for the $k$-th layer. A node $v$'s final representation $\bm{h}_v^{(K)} \in \mathbb{R}^{|\mathcal{Y}|}$ can capture the structural information of all nodes within $v$'s $K$-hop neighbors. Moreover, the final node representations of training nodes are used to train the node classifier. Specifically, let $\Theta = \{\bm{W}^{(1)}, \cdots, \bm{W}^{(K)} \} $ be the model parameters and $v$'s output be $ f_{\Theta}(\bm{a}_v) = \textrm{softmax}(\bm{h}_v^{(K)}) \in \mathbb{R}^{|\mathcal{Y}|}$, where $f_{\Theta}(\bm{a}_v)_{y}$ indicates the probability of node $v$ being class $y$\footnote{Note that the prediction also depends on $v$'s node feature $\bm{x}_v$ and the whole graph $G$. We omit $\bm{x}_v$ and  $G$ for notation simplicity.}. Then, $\Theta$ are learnt by minimizing the cross-entropy loss on the outputs of the training nodes $\mathcal{V}_L$, i.e.,  
{
\begin{align}
\small
    \label{obj:gcn}
    \Theta^* = \arg \min_{\Theta} - \sum_{v \in \mathcal{V}_L} 
    \ln f_{\Theta} (\bm{a}_v)_{y}.
\end{align}
}%
With the learnt $\Theta^*$, we can predict the label for each unlabeled nodes $u \in \mathcal{V} \setminus \mathcal{V}_L$ as $\hat{y}_u = \arg \max_{y} \, f_{\Theta^*}(\bm{a}_u)_{y}$.

\subsection{Threat model}
\noindent \textbf{Attacker's knowledge.} 
The considered black-box attack setting in this paper implies that an attacker does not know the internal configurations (i.e., the learned parameters) of the targeted GNN model. For a target node $v \in\mathcal{V} $, the only information the attacker knows about the GNN model is the predictions $f_{\Theta^*}(\bm{a}_v)$ (i.e., output logits) via querying the GNN model $f_{\Theta^*}$. Moreover, we also reasonably assume that the attacker knows her neighbors, i.e., the adjacency vector $\bm{a}_v$\footnote{For graph classification, we assume attackers know the input graph.}. In practice, the attacker naturally knows the neighbors of his controlled node. Taking social network as an instance, an attacker controls a malicious user, and this malicious user definitely knows his (non)neighbors in the social network. Note that the compared black-box RL-S2V attack \cite{dai2018adversarial} also requires that an attacker's target node knows his neighbors.

\noindent \textbf{Attacker's capability.} We consider that the attacker can modify the connection status (e.g., injecting new fake edges or removing the existing edges) between the target node $v$ and other nodes in the graph. In practice, it also incurs different costs for the attacker to manipulate different edges. The attacker's budget of manipulating edges is often limited, and we denote by $C$ the cost budget. We also constrain that the number of edges to be manipulated is bounded by $B$. 

\noindent \textbf{Attacker's goal.} 
Based on the attacker's knowledge and capability, an attacker's goal is to fool a targeted GNN, i.e., making her target node $v$'s predicted label different from the true label $y_{v}$, by perturbing her adjacency vector $\bm{a}_v$ with the cost budget $C$ and allowed number of perturbed edges $B$.

Our threat model requires that an attacker knows the confidence scores (as many existing attacks to DNN models). Although it is stronger than the threat model that an attacker only needs to know the hard label, we also highlight that this is the first optimization-based attack that targets discrete graph structure perturbation, where this problem itself is rather challenging. We will leave addressing the attack with hard labels as the query feedback in future work. 

\subsection{Problem formulation}
Given the target node $v$, label $y_v$, and adjacency vector $\bm{a}_v$, an attacker aims to modify the connection status related to the target node $v$ such that the targeted GNN misclassifies $v$. Let $\bm{s}_v \in\{0,1\}^{N}$ be the adversarial structure perturbation on $v$, where $s_{vu}=1$ means the connection status between the nodes $v$ and $u$ is changed, and $s_{vu}=0$, otherwise. Then, we define the perturbed adjacency vector for $v$ as $\bm{a}_v \oplus \bm{s}_v$, where $\oplus$ is the XOR operator between two binary vectors. Moreover, we denote $\bm{c}_v \in\mathbb{R}^{N}$ as the cost vector associated with $v$, i.e., $c_{vu}$ is the cost to modify the connection status between $v$ and $u$. In the focused black-box setting, we consider the untargeted attack. Let $L(\bm{a}_v)$ be the loss function for the targeted node $v$ without attack. With the adversarial perturbation $\bm{s}_v$, we have the attack loss as $L(\bm{a}_v \oplus \bm{s}_v)$. In this paper, we use the CW attack loss function \cite{carlini2017towards} with $\kappa$ attack confidence. Specifically, it is defined as follows:
\begin{equation}\label{eq:attack_loss}
\small
   L(x) = \max\{f_{\Theta^*}(x)_{{y}} - \max_{ \hat{y} \ne y}\{f_{\Theta^*}(x)_{\hat{y}}\},-\kappa\}. 
\end{equation}
Finally, our problem of the structure perturbation-based black-box attack to GNN can be formulated as
{
\footnotesize
\begin{equation}
\min_{\bm{s}_v}  L(\bm{a}_v \oplus \bm{s}_v),  \textrm{ s.t.}, 
\bm{1}^T\bm{s}_v \le B, 
\bm{c}^T\bm{s}_v \le C, 
\bm{s}_v\in \{0,1\}^N,
\label{TAGNN}
\end{equation}
}
where the first constraint means the number of edges to be perturbed is no more than $B$ and the second constraint means the total costs of the perturbation are no more than $C$. 
\vspace{-2mm}
\begin{lemma}\label{le:np}
Our problem in Eq.~(\ref{TAGNN}) is NP-hard.
\vspace{-2.5mm}
\begin{proof}
Our problem in Eq.~(\ref{TAGNN}) is a combinatorial optimization problem, actually a type of knapsack problem, which is a classical NP-hard problem. 
\end{proof}
\end{lemma}
\vspace{-2.5mm}

Lemma \ref{le:np} implies that it is difficult to calculate the optimal perturbation vector $\bm{s}_v^*$ within polynomial time under large graphs  (i.e., $\bm{s}_v$ has large dimension). To this end, we aim to design an approximation algorithm to derive sub-optimal solution. One algorithm is to relax the combinatorial binary constraint $\bm{s}_v\in \{0,1\}^N$ into convex hull $\bm{s}_v\in[0,1]^N$ and obtain a continuous optimization problem. Let $\bm{\hat{s}}_v$ be the solution of the continuous optimization problem. We can derive the sub-optimal solution for the original problem in Eq.~(\ref{TAGNN}) by rounding $\bm{\hat{s}}_v$ into combinatorial space $\{0,1\}^N$ using randomization sampling like Bernoulli sampling \cite{xu2019topology}. Then, we have the following lemma to characterize the relation between $\bm{s}_v$ and $\bm{\hat{s}}_v$ in expectation:
\vspace{-2mm}
\begin{lemma}
When sampling $\bm{s}_v$ element-wise in Bernoulli distribution using the probability from the relaxed vector $\bm{\hat{s}}_v\in[0,1]^N$, then the expectation of $\bm{s}_v$ is $\bm{\hat{s}}_v$, i.e., the condition $\mathbb{E}[\bm{s}_v] = \bm{\hat{s}}_v$ holds.
\vspace{-3.2mm}
\begin{proof}
This lemma holds due to the fact that a random variable $\bm{X}$ subject to Bernoulli distribution on support $\{0,1\}$ takes its probability as expectation, i.e., $\mathbb{E}[\bm{X}] = \mathbb{P}[\bm{X}]$. Applying this fact elements in $\bm{s}_v$, we can prove this lemma.
\end{proof}
\end{lemma}
\vspace{-3.2mm}

Conventionally, to solve our relaxed continuous optimization problem, we can apply the PGD approach by running gradient updates projected onto the feasible domain within several steps. However, PGD requires gradient information to be available. In our black-box attack setting, only the prediction result is available (by querying the GNN) instead of the exact gradient. Thus, the attack problem becomes how to estimate the gradient such that the PGD method can still be applied. It is shown that zeroth-order optimization (short for ZOO\footnote{Note that ZOO is a perfect benchmark in our setting. First, only ZOO approximates the gradient directly through queries. Second, ZOO is the only method that also has a regret bound. Hence, we can compare with ZOO in terms of both theoretical results and empirical attack performance.}) can be used to estimate the gradient \cite{chen2017zoo,ilyas2018prior,liu2018zeroth}. However, ZOO suffers from a low convergence rate and high query overhead due to necessarily exploring all edges to estimate the gradient per round. We aim to estimate the unknown gradient by controlling the exploration-exploitation tradeoff via bandit methods. Reinforcement learning (RL) can also control the exploration-exploitation tradeoff. However, RL-based attack, i.e., RL-S2V~\cite{dai2018adversarial}, is naturally heuristic. Our attack can address both issues in ZOO-based and RL-based attacks. Specifically, our attack has theoretical guarantees and better attack performance than ZOO-based and RL-based attacks (See  Section~\ref{eval}).  Next, we reformulate our attack problem as a bandit optimization problem and then propose a solution to it.

\subsection{Reformulating our attack as a bandit problem}
When the attacker selects a perturbation vector $\bm{s}_v$ and uses the perturbed adjacent vector $\bm{a}_v \oplus \bm{s}_v$ to query the GNN, the GNN returns the prediction $f_{\Theta^*}(\bm{a}_v \oplus \bm{s}_v)$. Thus, the objective $L(\bm{a}_v \oplus \bm{s}_v)$ is revealed based on Eq. (\ref{eq:attack_loss}), which can be seen as a bandit feedback (i.e., reward) for the selected perturbation vector $\bm{s}_v$ (i.e., an arm). Under the bandit feedback, the attacker wants to maximize the cumulative rewards. Note that since the attacker does not know the optimal arm $\bm{s}_v^*$ in each round, it will incur a regret, i.e., the difference of the maximum reward under the optimal arm $\bm{s}_v^*$ in hindsight and the reward of the attacker's attack algorithm. Then, the attacker's goal is to minimize the cumulative regrets. Let $\mathtt{Reg}(T)$ be the cumulative regrets in $T$ rounds, and $\bm{s}_v^t$ be the perturbation vector selected at round $t$, then the cumulative regrets $\mathtt{Reg}(T)$ can be calculated as $\mathtt{Reg}(T) = \mathbb{E}[\sum_{t=1}^TL(\bm{s}_v^t)] - TL({\bm{s}_v^*})$. In bandit optimization, it is important to design an arm selection algorithm with \emph{sublinear} regret (i.e., $\mathtt{Reg}(T) = o(T)$). This is because the selected arm $\bm{s}_v$ at round $T$ is asymptotically optimal when $T$ is sufficiently large (i.e., $\lim_{T\to\infty}\frac{\mathtt{Reg}(T)}{T} = 0$). 

\section{Structure Perturbation-based Black-Box Attacks to GNNs via Bandits}
Here, we design an online attack to GNN based on bandits optimization and show its sublinear regret in next section.

First, we relax the binary perturbation vector $\bm{s}_v\in \{0,1\}^N$ into a continuous convex hull $\bm{\hat{s}}_v\in[0,1]^N$. In this case, we can define the arm set $\mathcal{W}$ as: $\mathcal{W} = \{\bm{\hat{s}}_v\in[0,1]^N| \, \bm{1}^T\bm{\hat{s}}_v \le B,\bm{c}^T\bm{\hat{s}}_v \le C\},$ where $\mathcal{W}$ is convex. Note that our bandit for the black-box setting is different from the traditional multi-armed bandits (MAB), as arm set $\mathcal{W}$ contains infinite perturbation vectors. Thus, the approaches like upper confidence bound (UCB) \cite{auer2010ucb} and Thompson Sampling \cite{russo2017tutorial} in MAB fail to solve our problem. Moreover, it is impossible to use the combinatorial bandits~\cite{chen2013combinatorial,wen2017online} to derive the optimal perturbation because they have to collect enough historical samples to calculate the mean of each perturbed edge. In particular, in the exploitation phase, they behave as ZOO \cite{chen2017zoo} and estimate a gradient with $N+1$ queries. In the exploration phase, they require an approximation algorithm to derive the suboptimal perturbation with the UCB values. However, to the best of knowledge, there is no such an efficient approximation algorithm in the context of structural perturbation attacks with theoretical guarantees. 

Next, we need to address how to efficiently decide the next arm (i.e., $\bm{\hat{s}}_v^{t+1}$) at the end of round $t$ such that the regret is minimized. We leverage online convex optimization (OCO) technique to derive the next arm $\bm{\hat{s}}_v^{t+1}$ at round $t+1$. We note that the loss function $L(\cdot)$ is often non-convex. However, we emphasize that the gradient descent used in OCO is still useful as it is challenging to derive the closed-form solution for non-convex functions. OCO approach requires gradient information at the selected arm to conduct gradient descent. In our black-box attack setting, the attacker only receives the bandit feedback for the selected arm $\bm{\hat{s}}_v^t$. To estimate the gradient for the black-box attack at the selected arm $\bm{\hat{s}}_v^t$, the attacker can only query the GNN to compute the gradient. We use \emph{one point gradient estimation} (OPGE)~\cite{granichin1989stochastic} technique to estimate the gradient due to its simplicity and effectiveness. The idea of OPGE is to find a vector in unit sphere $\mathcal{S} = \{\bm{u}\in\mathbb{R}^{N}|~||\bm{u}||_2 = 1\}$ such that its direction has small intersection angle with the gradient (i.e., $\bm{u}$ is a good approximation of the gradient). To this end, we uniformly sample a unit vector from $\mathcal{S}$ and derive an approximate gradient in the following lemma.
\vspace{-2mm}
\begin{lemma}\label{le:smoothL}
For a unit vector $\bm{u}$ uniformly sampled from the unit sphere $\mathcal{S}$ and a sufficient small $\delta>0$, we can estimate gradient as $\nabla\hat{L}(\bm{\hat{s}}_v) = \mathbb{E}_{\bm{u}\in\mathcal{S}}[\frac{N}{\delta}L(\bm{\hat{s}}_v+\delta \bm{u})\bm{u}]$.
\end{lemma}

\begin{algorithm}[!t]
\footnotesize
    \renewcommand{\algorithmicrequire}{\textbf{Input:}}
    \renewcommand{\algorithmicensure}{\textbf{Output:}}
   \caption{Black-box attack to GNN for node classification via OCO with bandit}
   \label{alg:bb_ocob}
   \begin{algorithmic}[1]
        \REQUIRE Target GNN model $f_\Theta^*$, target node $v$, maximal \#perturbed edges $B$, cost budget $C$, PGD step $\eta$, $\delta>0$, $\alpha\in[0,1]$, query number $T$
        \ENSURE Perturbed vector $\bm{s}_v$
        \STATE Initialize: $\bm{v}^1 = \bm{0} \in \mathcal{W} = \{\bm{\hat{s}}_v\in[0,1]^N| \, \bm{1}^T\bm{\hat{s}}_v \le B,\bm{c}^T\bm{\hat{s}}_v \le C\}$
        \FOR {$t = 1$ \textbf{to} $T$}
         \STATE Attacker randomly chooses a unit vector $\bm{u}^t$ from the unit sphere $\mathcal{S}$.
         
         \STATE Attacker determines a perturbation $\bm{\hat{s}}_v^t  = \bm{v}^t + \delta \bm{u}^t$.                  
         
         \STATE Attacker converts $\bm{\hat{s}}_v^t$ to be binary $\bm{s}_v^t$ by setting top-$B$ values in $\bm{\hat{s}}_v^t$ to be 1 and others to be 0.         
         
         \STATE Attacker queries the GNN model $f_\Theta^*$ with $\bm{{s}}_v^t$ to obtain the predictions and CW loss $L(\bm{{s}}_v^t)$ using Eq. (\ref{eq:attack_loss}).
         \STATE Attacker conducts PGD and updates:
            $
             \bm{v}^{t+1} = \prod_{(1-\alpha)\mathcal{W}}(\bm{v}^t - \eta \bm{\hat{g}}), \, \bm{\hat{g}} = \frac{N}{\delta}L(\bm{{s}}_v^t)\bm{u}^t. 
            $
           
        \ENDFOR
        \STATE \textbf{return} $\bm{s}_v^T$
    \end{algorithmic}
\end{algorithm}
\setlength{\textfloatsep}{3mm}

The details of our attack algorithm are shown in Algorithm \ref{alg:bb_ocob}. The inputs of our algorithm include the targeted GNN model $f_\Theta^*$, target node $v$, maximal number of perturbed edges $B$, cost budget $C$, PGD step $\eta$, a small $\delta>0$, projection scale $\alpha\in[0,1]$, and query number $T$. The output is a perturbed vector $\bm{s}_v$ for the target node $v$ after $T$ rounds. In line 1, we set $\bm{0}$ as the initial prior vector $\bm{v}^1$. In line 2--8, we calculate a sub-optimal perturbed vector to attack the targeted GNN based on OCO with bandit feedback. At round $t$, we randomly select a unit vector $\bm{u}^t$ from the unit sphere $\mathcal{S}$ as a stochastic gradient in line 3. In line 4, we derive a relaxed perturbed vector $\bm{\hat{s}}_v^t$ by updating the prior vector $\bm{v}^t$ according to the selected stochastic gradient $\bm{u}^t$. In line 5, we \emph{convert $\bm{\hat{s}}_v^t$ to binary $\bm{{s}}_v^t$  by setting its top-$B$ nonzero values (which corresponds to the entries in $\bm{\hat{s}}_v^t$ with the $B$ largest nonzero probabilities) to be 1 (thus perturbing at most $B$ edges) and the remaining values to be 0}. In line 6, we query the GNN model $f_\Theta^*$ with $\bm{{s}}_v^t$ and obtain a loss feedback $L(\bm{{s}}_v^t)$. In line 7, we conduct PGD on the arm set $\mathcal{W}$ to update the $\bm{v}^{t+1}$ for round $t+1$. Finally, after $T$ queries, we obtain the perturbed vector $\bm{s}_v^T$ for the target node $v$.

\vspace{-6mm}
\subsection{Extending our attack for graph classification}
\vspace{-3mm}
Our proposed attack against node classification can be naturally extended to attack GNN models for graph classification with a small effort of modifications. In a graph classification model, its input is an adjacent matrix of a graph and its output is the label of the graph. In node classification, we aim to perturb the adjacent vector of a target node, while in graph classification, we perturb the adjacent matrix. Let $\bm{A} \in\{0,1\}^{N\times N}$ be the adjacent matrix of a graph with $N$ nodes. Moreover, let $\bm{S} \in\{0,1\}^{N\times N}$  be a perturbation matrix,  where $\bm{S}_{ij} = 1$ if the connection status between the edge $(i,j)$ is modified, and $\bm{S}_{ij} = 0$, otherwise. To perform our attack against graph classification models, we only need to flatten the matrix $\bm{S}$ into a vector $\bm{s}$ and feed it as an input to our attack algorithm. After obtaining the perturbed $\bm{s}$, we can reshape it to a perturbed adjacent matrix.

\section{Main Results}
In this section, we analyze the regret bound of our attack algorithm where we assume the loss function is convex. We note that it is an interesting future work to generalize our analysis to non-convex loss functions. We first present the following assumptions and lemmas. 

\begin{assumption}\label{ass:Lips}
There exists a Lipschitz constant $C_L$ such that the following inequality holds for any $\bm{u}$ and $\bm{v}$,
\vspace{-2mm}
{
\begin{align}\label{eq:Lip}
\small
|L(\bm{u}) - L(\bm{v})| \le C_L ||\bm{u} - \bm{v}||_2.
\end{align}
}
\end{assumption}

\begin{lemma}\label{le:relaxreg}
For continuous $\bm{\hat{s}}_v^t$ and the rounded binary $\bm{s}_v$, the instant regret is bounded as: 
\vspace{-2mm}
{
\begin{align}
\small
\mathbb{E}[|L(\bm{s}_v) - L(\bm{\hat{s}}_v^t)|] \le C_LN^{3/4}\sqrt{1+\frac{\eta}{\delta}}.
\end{align}  
}
\end{lemma}
\vspace{-3mm}
Line 7 in Algorithm \ref{alg:bb_ocob} can be seen as the stochastic gradient decent on $L(\bm{\hat{s}}_v+\delta \bm{u})$. Thus, we have the following lemma to characterize its regret bound.
\vspace{-2mm}
\begin{lemma}[\cite{flaxman2004online}]\label{le:sgdregret}
Suppose that the arm set $\mathcal{W}$ satisfies $\mathcal{W}\subseteq R\mathbb{B}$ for given radius $R>0$, where $\mathbb{B}$ is unit ball, i.e., $\mathbb{B} = \{\bm{u}\in\mathbb{R}^{N}: ||\bm{u}||_2 \le1 \}$. When loss function $L(\cdot)$ is convex, the cumulative regret for relaxed continuous variables are bounded as follows:
\vspace{-2mm}
{
\begin{equation}
\small
\mathbb{E}[\sum_{t=1}^TL(\bm{\hat{s}}_v^t)] - TL(\bm{\hat{s}}_v^*) \le C_LR\sqrt{T},
\end{equation}
}
where $\bm{\hat{s}}_v^t$ is continuous variable at round $t$ and $\bm{\hat{s}}_v^*$ is optimal continuous solution of our relaxed optimization problem. 
\end{lemma}
\vspace{-3mm}

Note that Lemma \ref{le:sgdregret} just captures the regret on the whole arm set $\mathcal{W}$. However, line 7 in Algorithm \ref{alg:bb_ocob} updates the arm $\bm{\hat{s}}_v^t$ by projecting onto set $(1-\alpha)\mathcal{W}$ for $0<\alpha<1$. The incurred regret by conducting $(1-\alpha)$-projection is captured by the following lemma.
\vspace{-2mm}
\begin{lemma}\label{le:scalepro}
For time horizon $T$, the incurred regret due to $(1-\alpha)$-projection is bounded as follows:
\vspace{-2mm}
\begin{equation}
\min_{\bm{w}\in(1-\alpha)\mathcal{W}}\sum_{t=1}^TL(\bm{w}) - TL(\bm{\hat{s}}_v^*) \le 2\alpha T.
\end{equation}
\end{lemma}
\vspace{-3mm}

Based on the above assumption and the lemmas, we have the following theorem on the regret bound of our attack: 
\begin{theorem}\label{th:regbound}
Under $T$ rounds attack span, our proposed attack algorithm incurs regret $\mathtt{Reg}(T)$, which is upper bounded by $\mathcal{O}(\sqrt{N}T^{3/4})$ with $T$ queries to the GNN.
\end{theorem}

\noindent \textbf{Remark.} Theorem \ref{th:regbound} not only demonstrates the sublinear regret our attack achieves, but also presents that our attack is dimension-efficient, i.e., $\mathcal{O}(\sqrt{N})$. It implies that our attack can be scalable to large graphs. Note that gradient-free ZOO \cite{chen2017zoo} has query complexity $\mathcal{O}(N)$. From the regret bound, we can see that our attack enjoys a better convergence rate $\mathcal{O}(1/T^{3/4})$ than ZOO~\cite{liu2018zeroth}, which has a  $\mathcal{O}(1/T^{1/2})$ convergence rate. Note also that Ilyas \emph{et al.} \cite{ilyas2018prior} proposed a bandit-based black-box attack to image classifiers, which can be adapted to solve our problem. However, their approach 1) does not provide theoretical results in terms of regret bound; and 2) is less efficient than ours due to requiring multiple gradient estimation in each iteration. 

\noindent {\bf Computational complexity.} Our attack requires 1 query per round and each query has time complexity $O(N)$. ZOO requires $2N$ queries per round and each query has time complexity $O(N)$---Its time complexity per round is $O(N^2)$. RL-S2V needs to trains an extra $Q$-network, which is used to perform the attack. During the attack, it has the same time complexity as our attack: 1 query per round and each round has a time complexity $O(N)$. These  analyses show our attack is more efficient than RL-S2V and ZOO.

\section{Related Work}
Recent attacks to GNNs mainly focus on white-box/gray box settings \cite{zugner2018adversarial,dai2018adversarial,zugner2019adversarial,xu2019topology,wang2020efficient,sun2020adversarial,chang2020restricted,zhang2020backdoor,ma2020towards}. For instance, Zugner~\cite{zugner2018adversarial} proposed the first attack, called Nettack, against GCN for node classification by perturbing graph structure or/and node features. Specifically, Nettack learns a surrogate linear model of GCN by defining a graph structure-preserving perturbation that constrains the difference between the node degree distributions of the graph before and after an attack. Xu et al.~\cite{xu2019topology} utilized the model gradient to generate perturbation on the topology of the graph. We study the black-box setting where gradient information is unknown to the attacker. A recent work~\cite{ma2020towards} performed black-box attacks to  GNNs. However, This work focus on perturbing the continuous node features, which does not fit our problem well and is less effective than discrete structure perturbation. In addition, the attack is implemented via a heuristic greedy algorithm, which has no theoretically guaranteed performance. Also, \cite{zugner2018adversarial} has shown that feature perturbation-based attacks to GNNs is significantly less effective than structure perturbation-based attacks (e.g., an attacker needs to perturb an average of 100 node features, in order to have a comparable performance by the attack that perturbs only 5 edges). Thus, we focus on the graph structure perturbation in this paper. Zang et al.~\cite{zang2020graph} studied the graph universal adversarial attacks where a set of anchor nodes is identified and their connection to the target node is flipped. However, how to select the optimal anchor nodes is not investigated. Some other works~\cite{waniek2018hiding,li2020adversarial} focus on attacking community detection and they are heuristic and orthogonal to our work. 
The work~\cite{mu2021hard} most close to ours designed a black-box attacks to GNNs for graph classification, which is based on gradient-free ZOO~\cite{chen2017zoo}. However, it also does not have theoretical guaranteed attack performance. 

Several black-box attacks \cite{ilyas2018black,ilyas2018prior,cheng2019improving} for non-graph classification models have been proposed. However, these methods cannot solve our problem because their attack problems are essentially continuous optimization problems. Note that \cite{ilyas2018prior} also used the bandit to formulate the black-box attack problem.  However, their work does not have a theoretical regret bound and is less efficient due to using multiple gradient estimations in each iteration.

\section{Experiments}
\label{eval}
\subsection{Experimental Setup}
\noindent {\bf Dataset description and GNN models.} In node classification experiments, we use three benchmark citation graphs, i.e., Cora, Citeseer, and Pubmed~\cite{sen2008collective}. In these graphs, each node represents a document and each edge indicates a citation between two documents. Each document treats the bag-of-words feature as the node feature vector, and has a label. We adopt the representative GCN~\cite{kipf2017semi} and SGC~\cite{wu2019simplifying} for node classification, and evaluate our attack against the two models. In graph classification, we use two benchmark superpixel graphs, i.e., MNIST and CIFAR10~\cite{dwivedi2020benchmarkgnns}, in computer vision. MNIST and CIFAR10 are classical image classification datasets. In our experiments, they are converted into graphs using the SLIC super-pixels~\cite{achanta2012slic}. Each node has the super-pixel coordinates as the feature and each super-pixel graph has a label. We adopt the representative GIN~\cite{xu2019how} for graph classification, and evaluate our attack against GIN. Table \ref{tab:exp} summarizes the basic statistics of these  datasets. 

\noindent {\bf Training nodes/graphs and target nodes/graphs.} We use the training nodes/graphs to train GNN models and use the target nodes/graphs to evaluate our attack against the trained models. Following existing works~\cite{kipf2017semi,zugner2018adversarial,dwivedi2020benchmarkgnns}, in citation graphs, we randomly sample 20 nodes from each class as the training nodes; sample 100 nodes that are correctly classified by each GNN model as the target nodes. In image graphs, we respectively use 55,000 graphs and 45,000 graphs in MNIST and CIFAR10 for training, and randomly sample 100 graphs correctly classified by the GIN model as the target graphs. 

\noindent {\bf Baselines.}  We compare our attack with two state-of-the-arts: \emph{RL-based attack} \cite{dai2018adversarial} and \emph{ZOO attack}~\cite{chen2017zoo}. Note that for the RL-based attack, we adjust the reward using the same CW attack loss like ours, and thus it has a fair comparison with our attack. We also choose \emph{random attack} for comparison, where we generate structure perturbations by randomly changing connection status between pairs of nodes. 

\noindent {\bf Cost simulation.} We specify the cost of modifying the connection state for each pair of nodes. Note that the costs could be application-dependent. W.l.o.g., we assume the costs for different node pairs are uniformly distributed among a certain interval (e.g., [1,5] in our experiments). Note that the equal cost can be seen as a special case of the uniform cost.

\noindent {\bf Parameter setting.} We set the hyperparameters in our attack algorithm as follows: $\eta = 10^{-4}$, $\delta = 10^{-6}$, and $\alpha = 0.7$ for attacking node classification methods, and $\eta = 10^{-1}$, $\delta = 10^{-3}$, and $\alpha = 0.6$ for attacking graph classification methods. Consider the different graph sizes, we set the default maximal number of perturbed edges $B$ as $5$ and $15$ on the three citation graphs, and on the two image graphs, respectively; and the default total costs $C$ is bounded by $25$ on the citation graphs and $75$ on the image graphs, respectively. In addition, the total number of queries $T=50$ by default. We also study the impact of these parameters in our experiments. We implement our algorithm in Python and conduct experiments using public source codes. All our experiments are run in a computer with Intel(R) Core(TM) i7-6700 CPU @3.4Hz processors with 4 cores, 32GB RAM, 1 TB disk space and 6G GPU. We run all experiments $30$ times and report the average result.

\noindent {\bf Evaluation metric.} We adopt the attack successful rate, i.e., a fraction of the total (i.e., 100) targeted nodes/graphs misclassified after our attack, as the metric to measure the effectiveness of our attack. We also use the number of queries ($T$) to measure the efficiency of our attack. Specifically, we count each PGD iteration in our Algorithm~\ref{alg:bb_ocob} as a query.  

\begin{table}[t]
\footnotesize
\centering
\addtolength{\tabcolsep}{-4pt}
\caption{Dataset statistics}\label{tab:exp}
\begin{tabular}{|l|l|l|l|l|l|}
 \hline
{\bf Datasets}& \#Graphs & \#Ave. Nodes & \#Ave. Edges & \#Classes & Task\\
\hline
{\bf Cora} & 1 & 2,708 & 5,429 & 7 & {\text{(1)}}\\
\hline
{\bf Citeseer} & 1 & 3,327 & 4,732 & 6 & {\text{(1)}}\\
\hline
{\bf Pubmed} & 1 & 19,717 & 44,338 & 3 & {\text{(1)}} \\
\hline
{\bf MNIST} & 70K &  70.57 & 564.53 & 10 & {\text{(2)}}\\
\hline
{\bf CIFAR10} &60K & 117.63 &  941.07 & 10 & {\text{(2)}}\\
 \hline
\end{tabular}
\\ {(1) \text{~Node classification,~} (2) \text{~Graph classification}}
\end{table}

\begin{figure*}[t]
\centering
\subfigure[Cora.]{\includegraphics[width=0.32\textwidth]{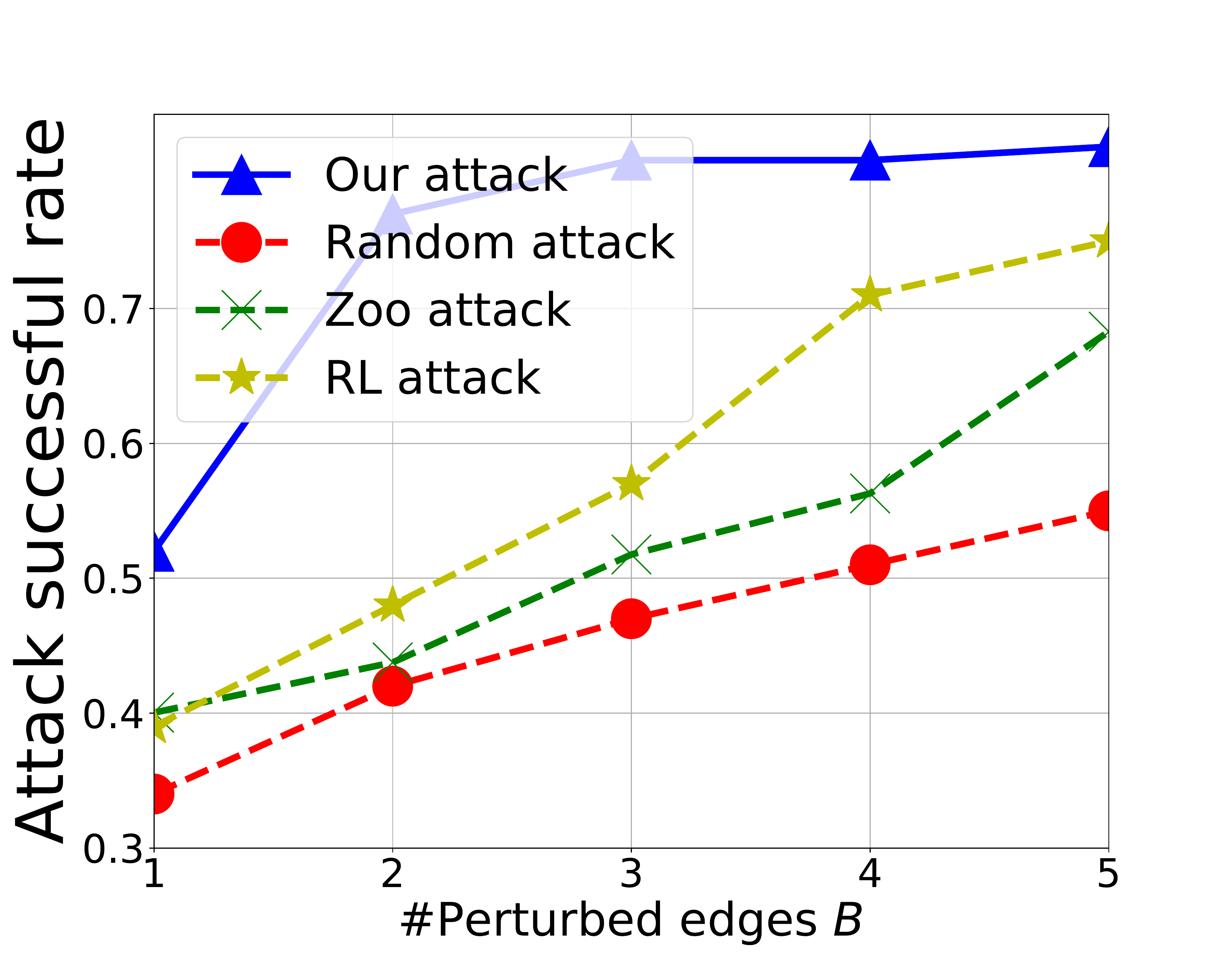}}
\subfigure[Citeseer.]{\includegraphics[width=0.32\textwidth]{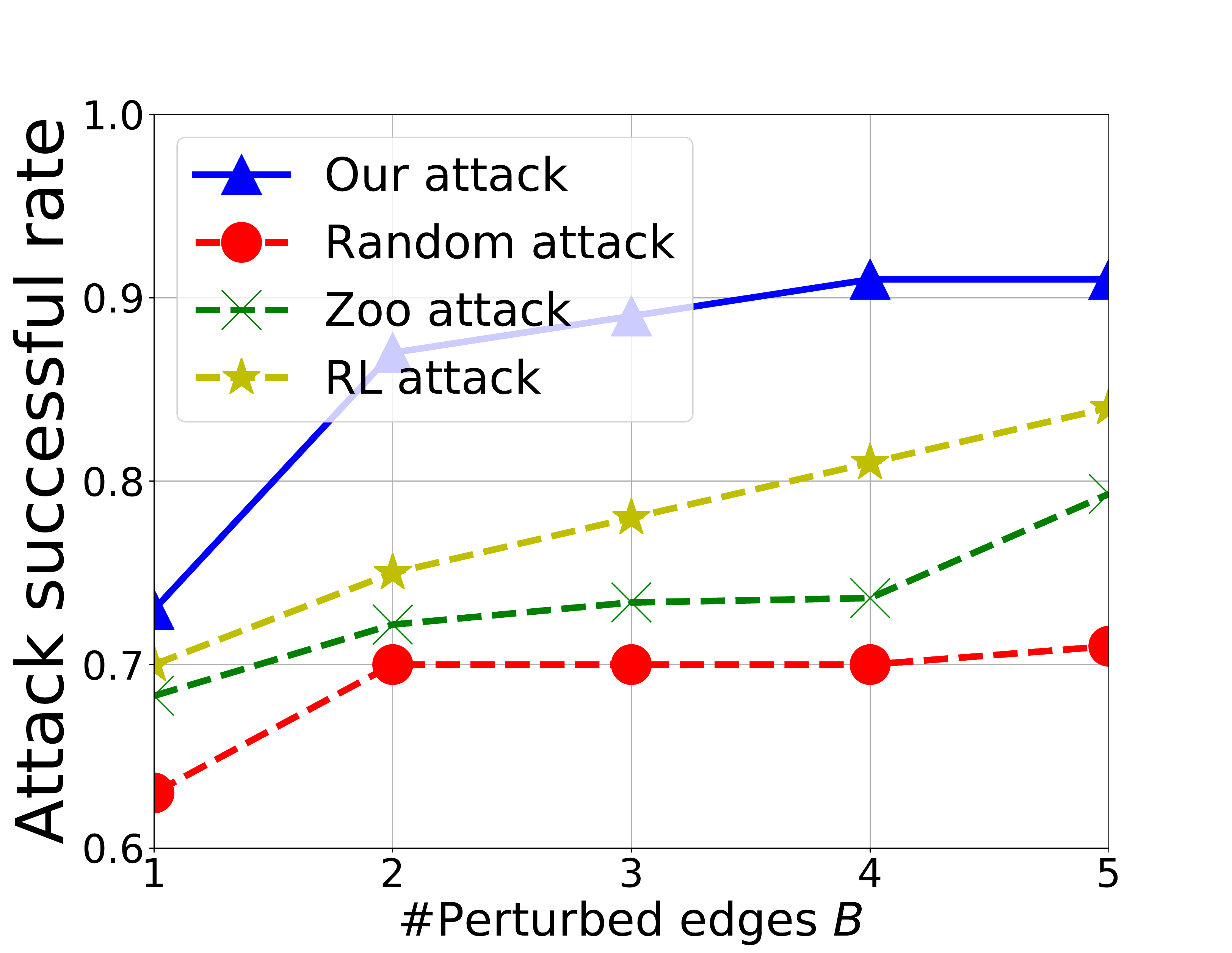}}
\subfigure[Pubmed.]{
\includegraphics[width=0.32\textwidth]{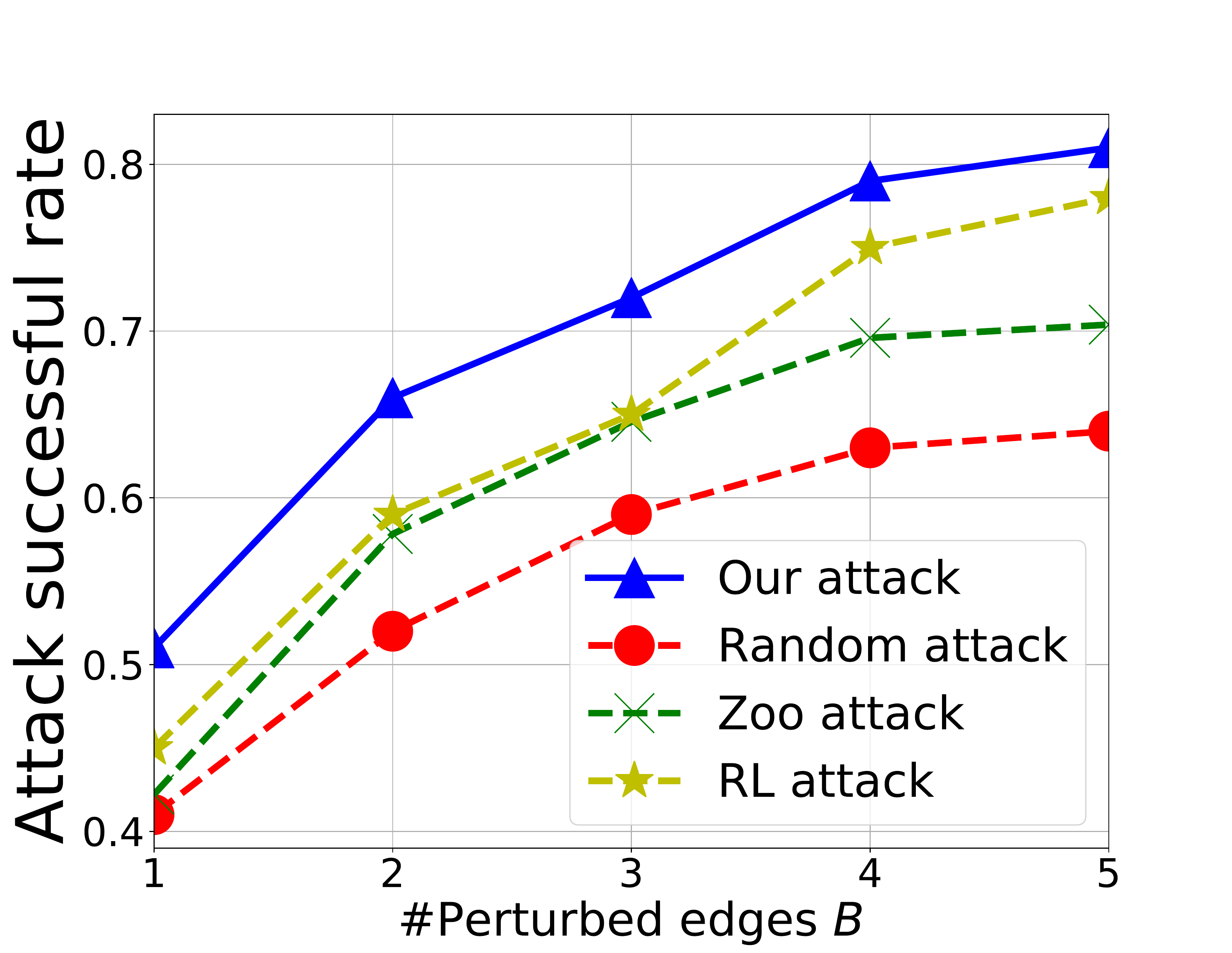}}
\caption{Attack successful rate vs. \#perturbed edges $B$ on GCN for node classification.} 
\label{fig:impact_edges_GCN}
\vspace{-5mm}
\end{figure*}

\begin{figure*}[t]
\centering
\subfigure[Cora.]{
\centering
\includegraphics[width=0.29\linewidth]{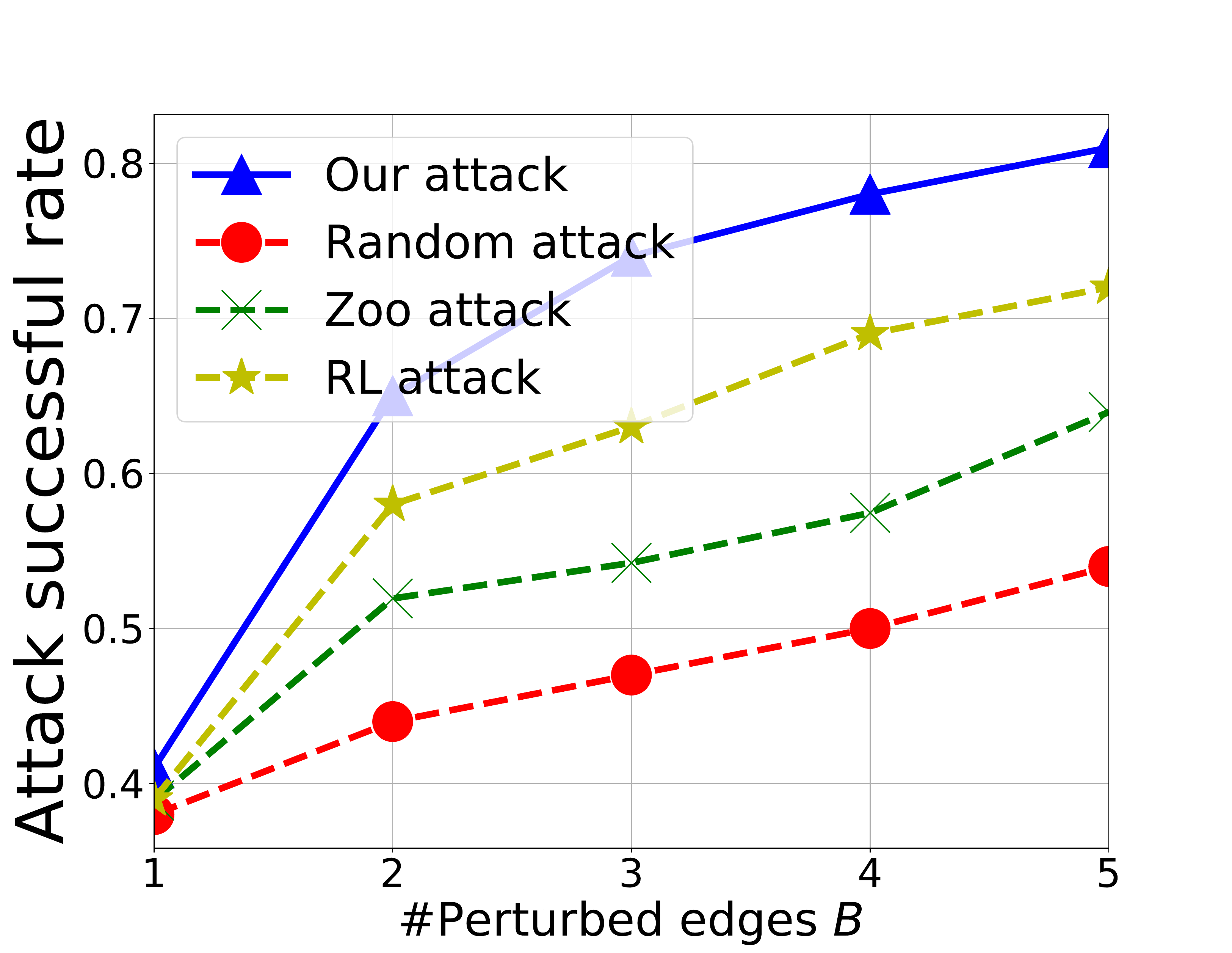}
}%
\subfigure[Citeseer.]{
\centering
\includegraphics[width=0.29\linewidth]{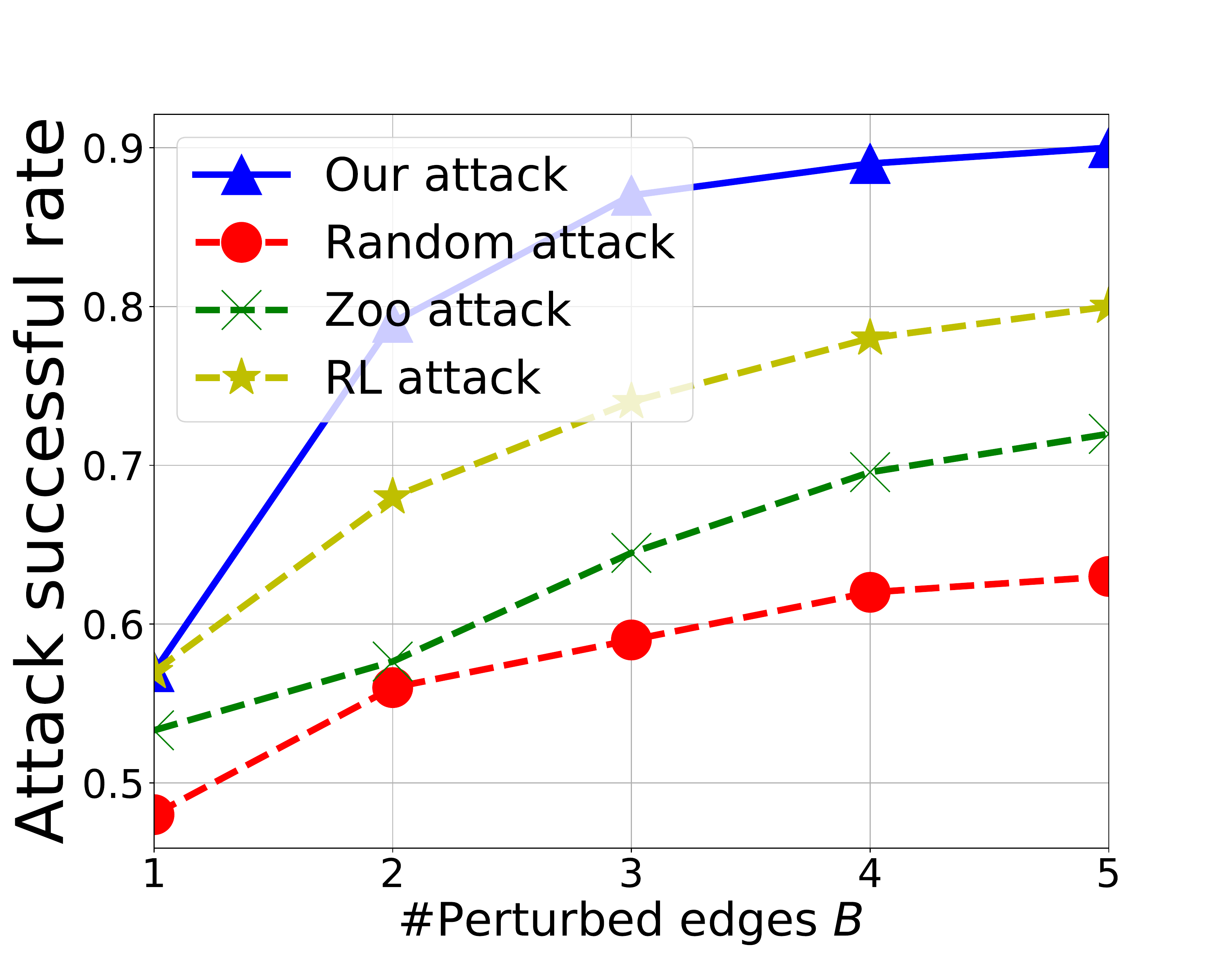}
}%
\subfigure[Pubmed.]{
\centering
\includegraphics[width=0.29\linewidth]{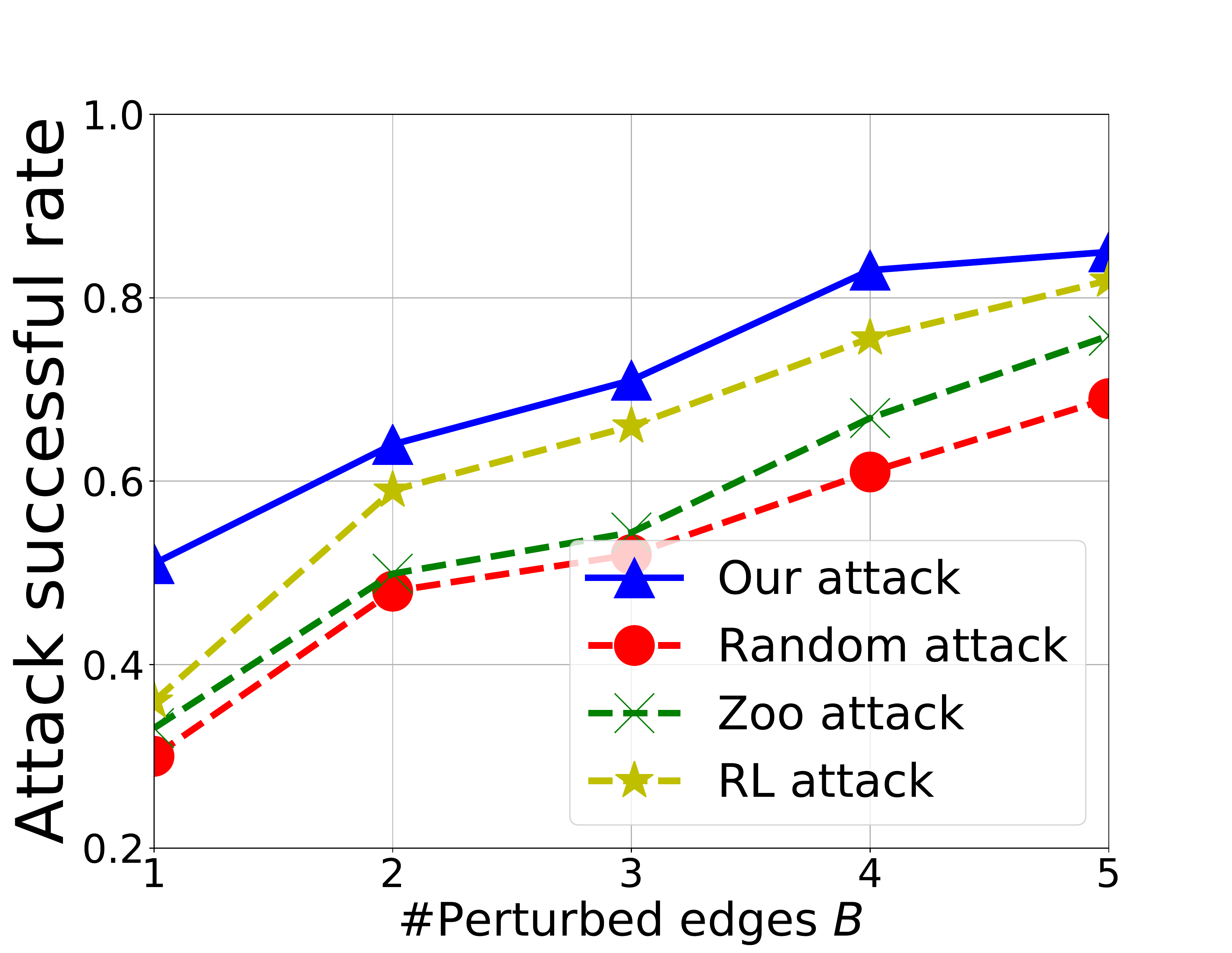}
}%
\centering
\caption{Attack successful rate vs. \#perturbed edges $B$ on SGC for node classification.} 
\label{fig:impact_edges_SGC}
\vspace{-5mm}
\end{figure*}

\begin{figure*}[!t]
\centering
\begin{minipage}[t]{0.48\linewidth}
\subfigure[MNIST.]{
\centering
\includegraphics[width=0.5\textwidth]{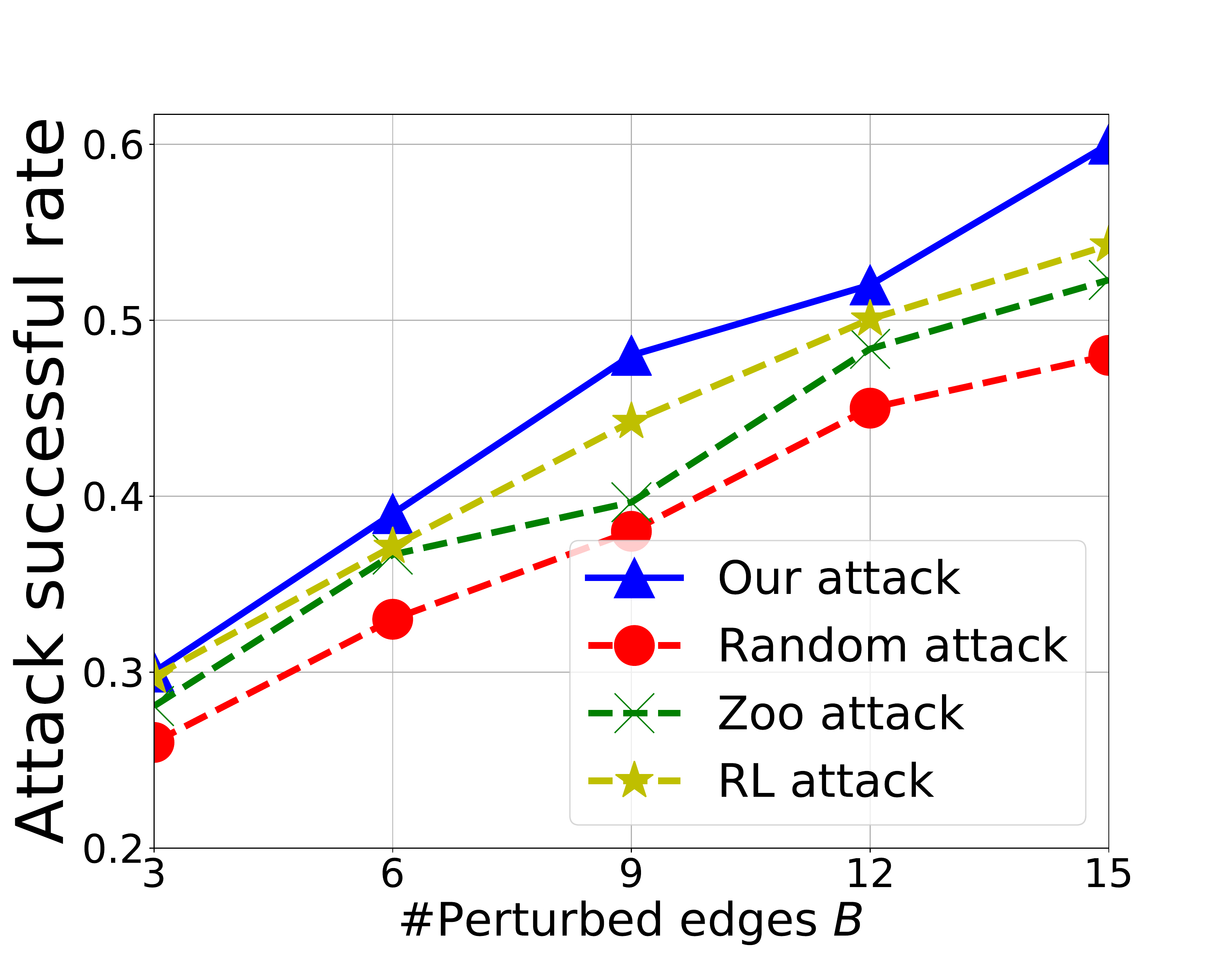}
}%
\subfigure[CIFAR10.]{
\centering
\includegraphics[width=0.5\textwidth]{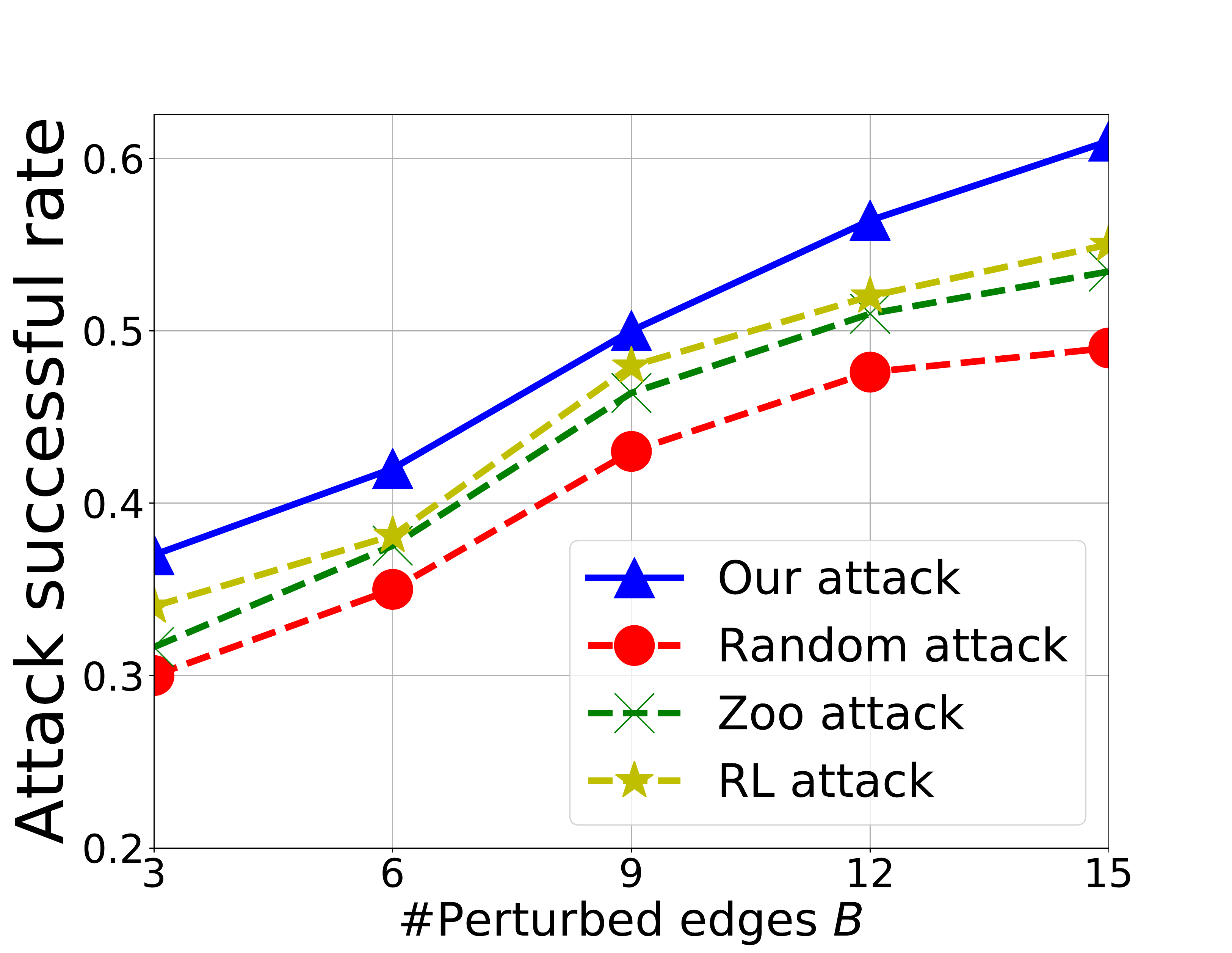}
}
\caption{Attack successful rate vs. \#perturbed edges $B$ on GIN for graph classification.}
\label{fig:impact_edges_GIN}
\end{minipage}%
\hfill
\begin{minipage}[t]{0.48\linewidth}
\subfigure[MNIST.]{
\centering
\includegraphics[width=0.5\textwidth]{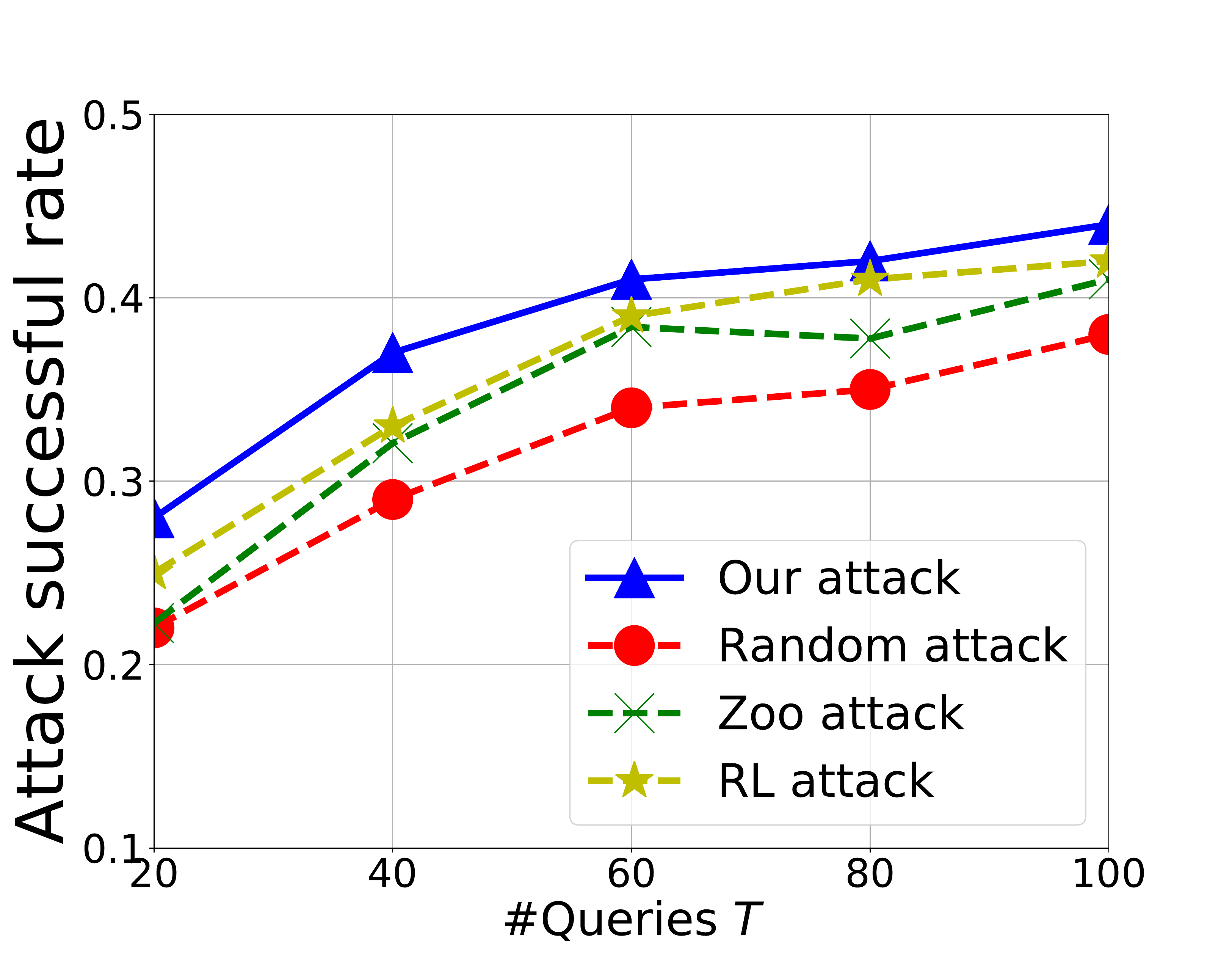}
}%
\subfigure[CIFAR10.]{
\centering
\includegraphics[width=0.5\textwidth]{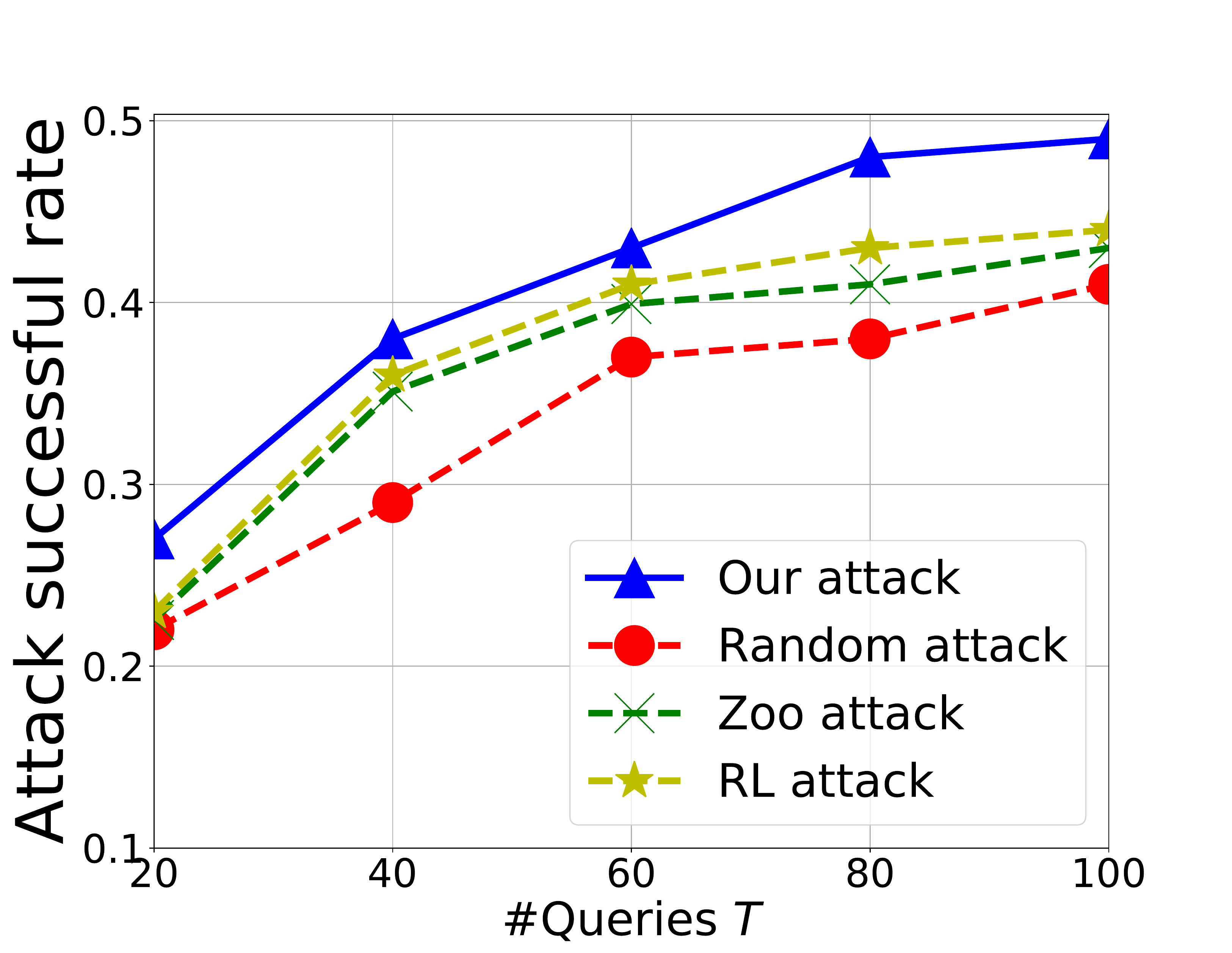}
}
\caption{Attack successful rate vs. \#queries $T$ on GIN for graph classification.}
\label{fig:impact_query_GIN}
\end{minipage}%
\vspace{-4mm}
\end{figure*}

\subsection{Experimental Results}
In this section, we provide a comprehensive evaluation of our black-box attack against GNNs for both node classification and graph classification. We aim to study our attack in terms of both effectiveness and efficiency. Our attack algorithm has three key factors: the number of perturbed edges $B$, total costs $C$, and the number of queries $T$. We will study the impact of these parameters one by one. 

\vspace{-6mm}
\paragraph{Impact of the number of perturbed edges.}
In this experiment, we separately fix the bounded total costs $C$ to be $25$ and $75$, and set the number of queries $T$ to be $50$ on attacking both node classification models and graph classification models. The results of attacking these models are shown in Figure~\ref{fig:impact_edges_GCN}, Figure~\ref{fig:impact_edges_SGC}, and Figure~\ref{fig:impact_edges_GIN}, respectively. 

We have the following observations: First, all attack approaches achieve a larger attack successful rate when the maximal number of perturbed edges $B$ increases. This is because a larger $B$ allows an adversary to have a better capability to perform the attack. Second, our attack significantly achieves a higher attack successful rate than the compared attacks in all GNNs models and graph datasets. For instance, when attacking GCN and SGC for node classification, our attack achieves the attack successful rate larger than $80\%$ (and even $90\%$), while the second-best RL attack achieves the attack successful rate at most $80\%$ across the three citation graphs. When attacking GIN for graph classification, our attack achieves $60\%$ attack successful rates, while the RL attack obtains less than $53\%$ attack successful rate in the two image graphs. All these results verify that the black-box bandit feedback in our algorithm is very useful to guide the selection of the optimal edges to be perturbed. In contrast, the heuristic RL attack is hard to do so. 

\begin{figure*}[!t]
\centering
\subfigure[Cora.]{
\begin{minipage}[t]{0.29\linewidth}
\centering
\includegraphics[width=\columnwidth]{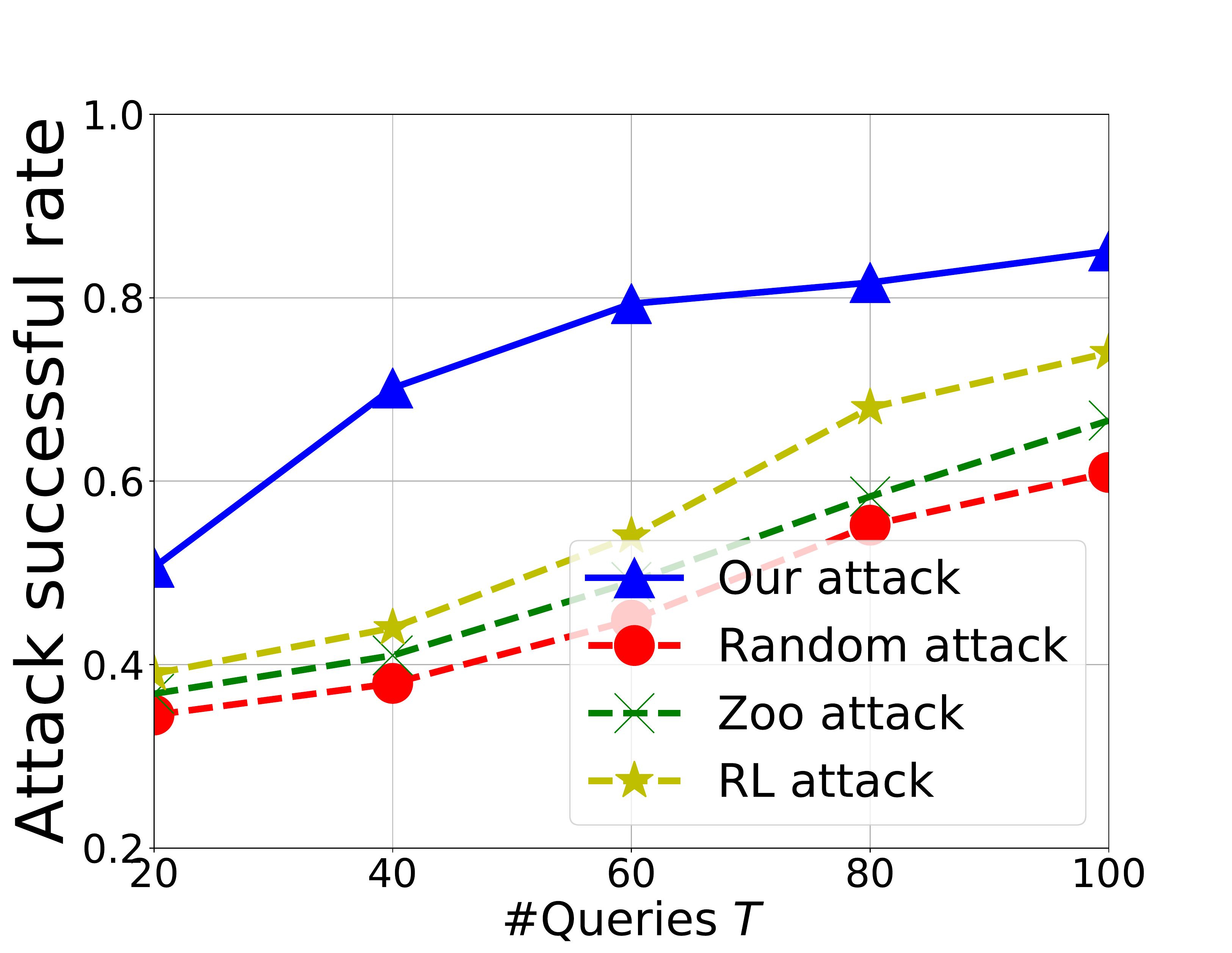}
\end{minipage}%
}%
\subfigure[Citeseer.]{
\begin{minipage}[t]{0.29\linewidth}
\centering
\includegraphics[width=\columnwidth]{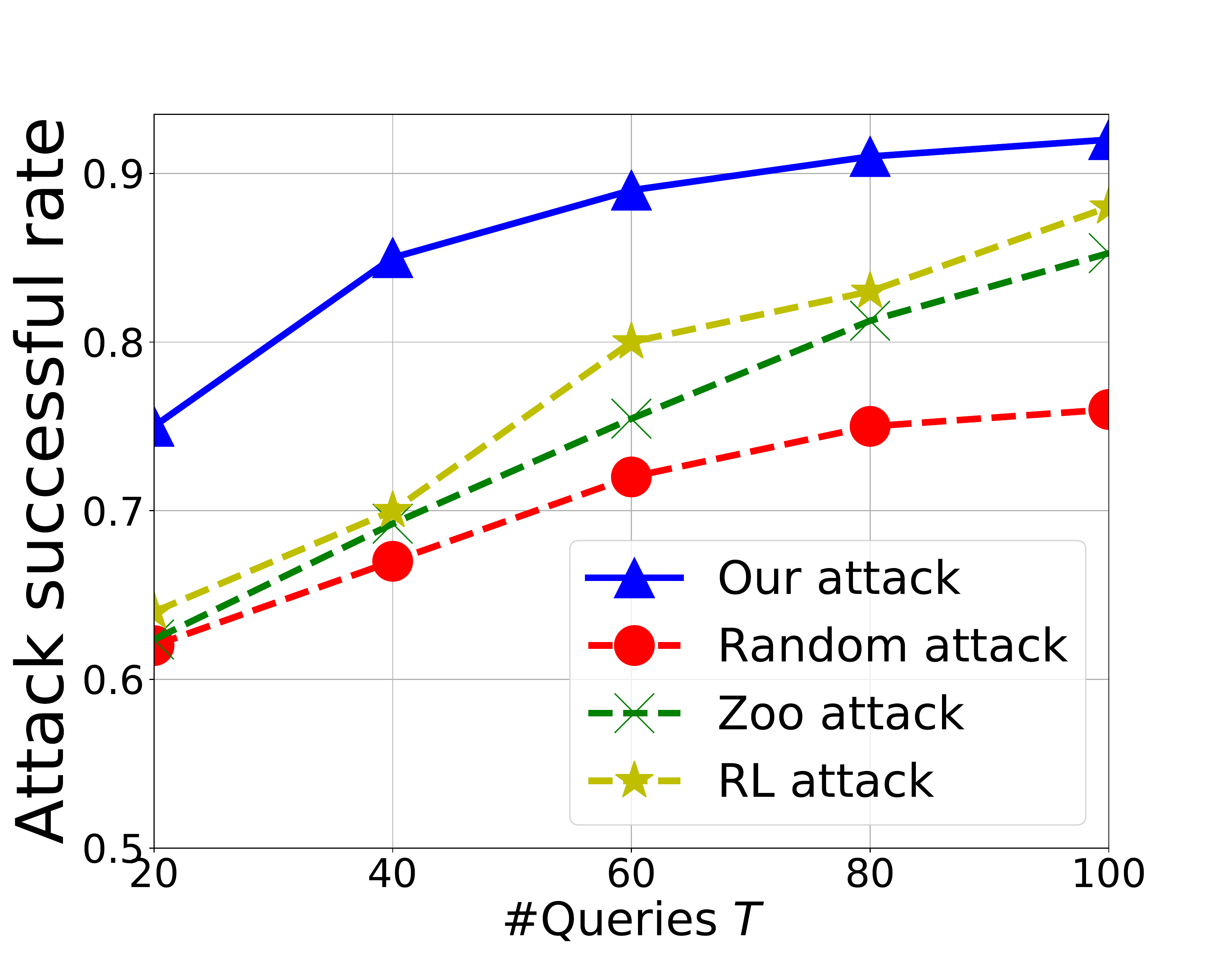}
\end{minipage}%
}%
\subfigure[Pubmed.]{
\begin{minipage}[t]{0.29\linewidth}
\centering
\includegraphics[width=\columnwidth]{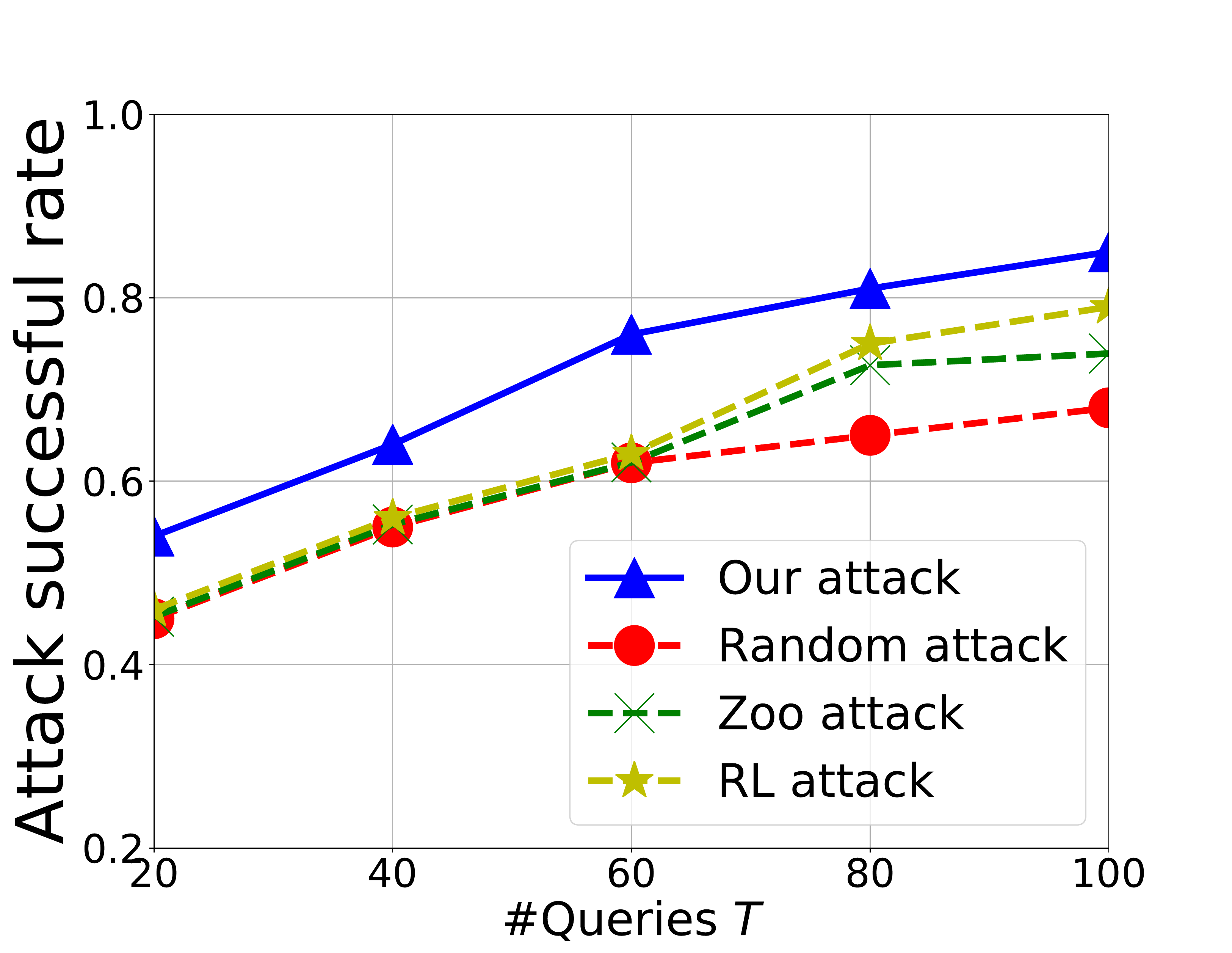}
\end{minipage}%
}%
\centering
\caption{Attack successful rate vs. \#queries $T$ on GCN for node classification.} 
\label{fig:impact_query_GCN}
\vspace{-4mm}
\end{figure*}

\begin{figure*}[!t]
\centering
\subfigure[Cora.]{
\begin{minipage}[t]{0.29\linewidth}
\centering
\includegraphics[width=\columnwidth]{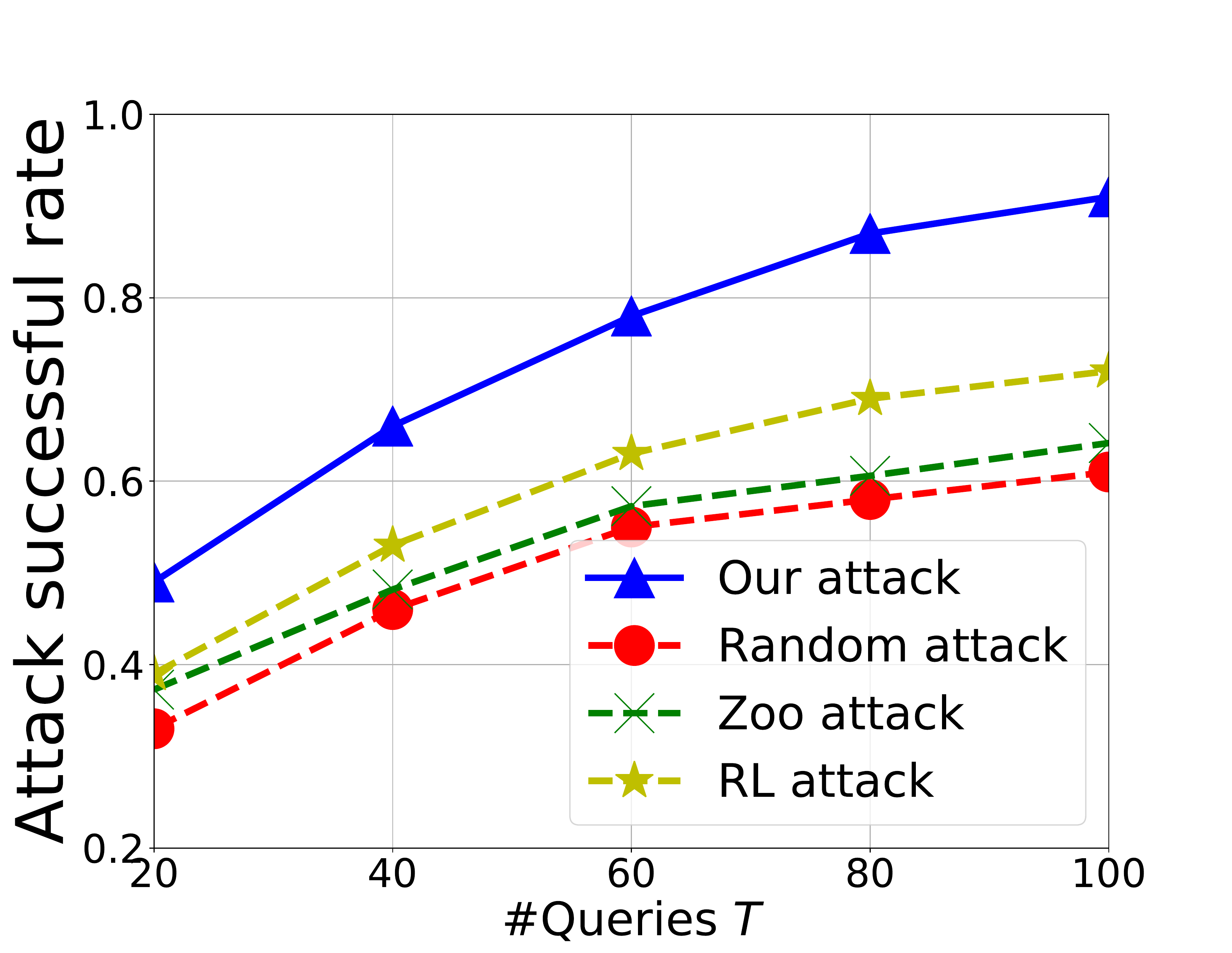}
\end{minipage}%
}%
\subfigure[Citeseer.]{
\begin{minipage}[t]{0.29\linewidth}
\centering
\includegraphics[width=\columnwidth]{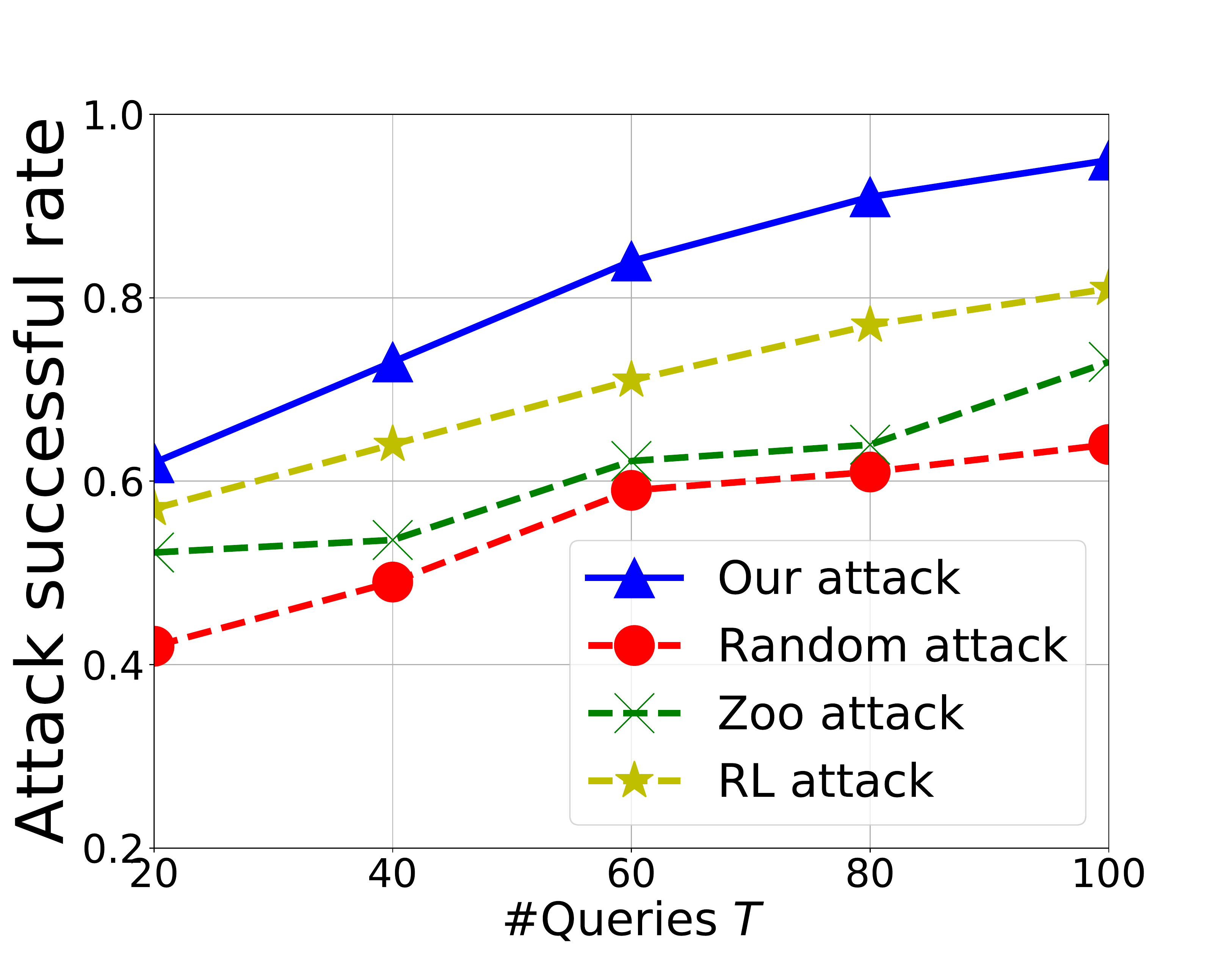}
\end{minipage}%
}%
\subfigure[Pubmed.]{
\begin{minipage}[t]{0.29\linewidth}
\centering
\includegraphics[width=\columnwidth]{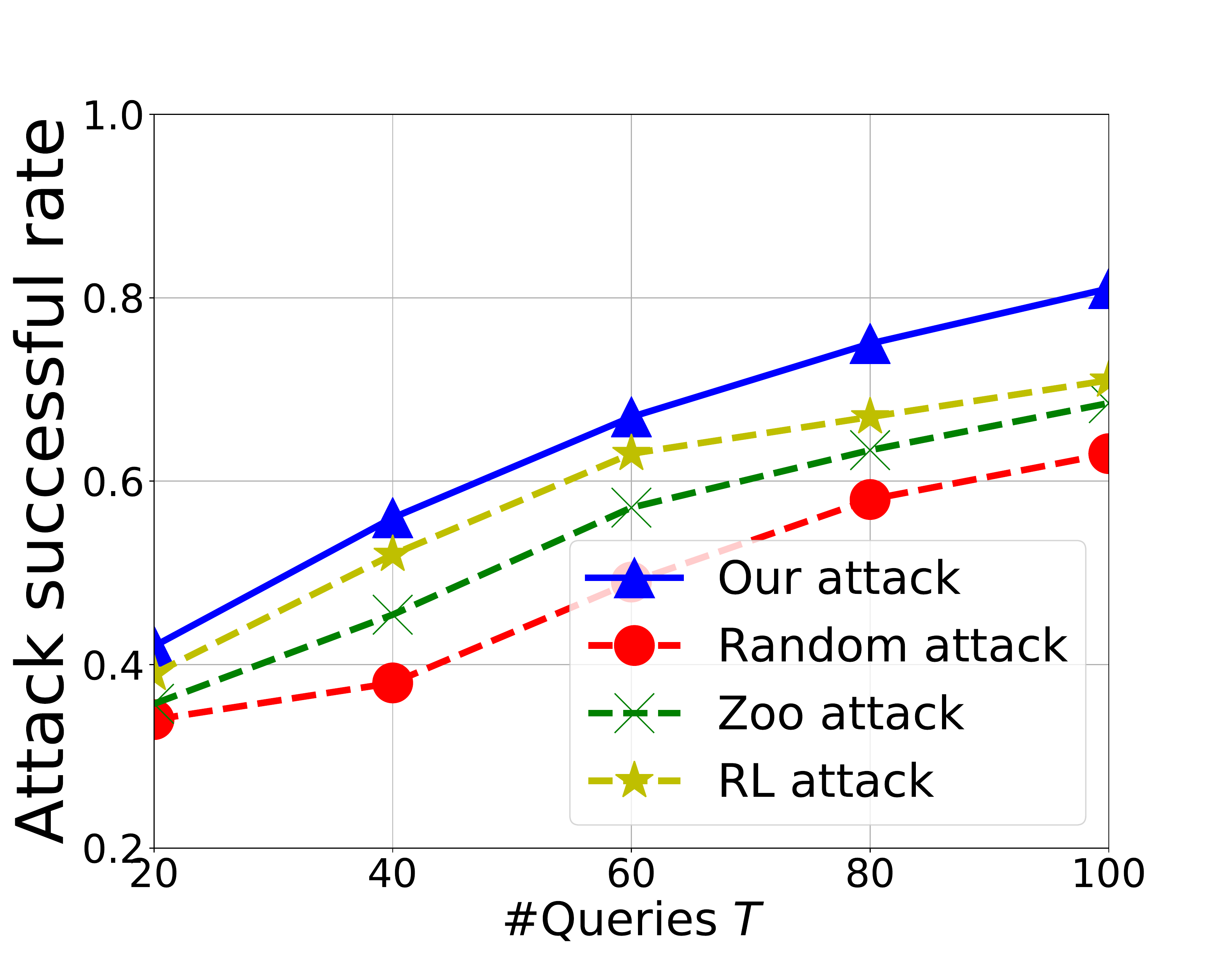}
\end{minipage}%
}%
\centering
\caption{Attack successful rate vs. \#queries $T$ on SGC for node classification.} 
\label{fig:impact_query_SGC}
\vspace{-4mm}
\end{figure*}

\vspace{-2mm}
\paragraph{Impact of the number of queries.}
In this experiment, we separately fix the bounded total costs $C$ to be $25$ and $75$, and the maximal number of perturbed edges $B$ to be $2$ and $5$ on attacking node classification models and graph classification models, respectively. The results of attacking GIN for graph classification and attacking GCN and SGC for node classification are shown in Figure~\ref{fig:impact_query_GIN}, Figure~\ref{fig:impact_query_GCN}, and Figure~\ref{fig:impact_query_SGC}, respectively. First, all attack approaches achieve a larger attack successful rate when the number of queries $T$ increases. This is because a larger $T$ allows an adversary to obtain more predictions via querying the GNN models and thus have a better capability to perform the attack. Second, our attack requires much fewer queries to achieve the same attack successful rate than the compared attacks. In particular, compared to our attack, RL attack requires 1.5-3x queries, when achieving comparable attack performance against node classification models and graph classification models, respectively. Third, given the same number of queries, our attack achieves $10\%-20\%$ higher successful rate than RL attack across all citation graphs and image graphs. Similarly, these results again verify that the black-box bandit feedback is really beneficial in designing our attack.  

\begin{figure*}[!t]
\centering
\subfigure[Cora.]{
\begin{minipage}[t]{0.29\linewidth}
\centering
\includegraphics[width=\columnwidth]{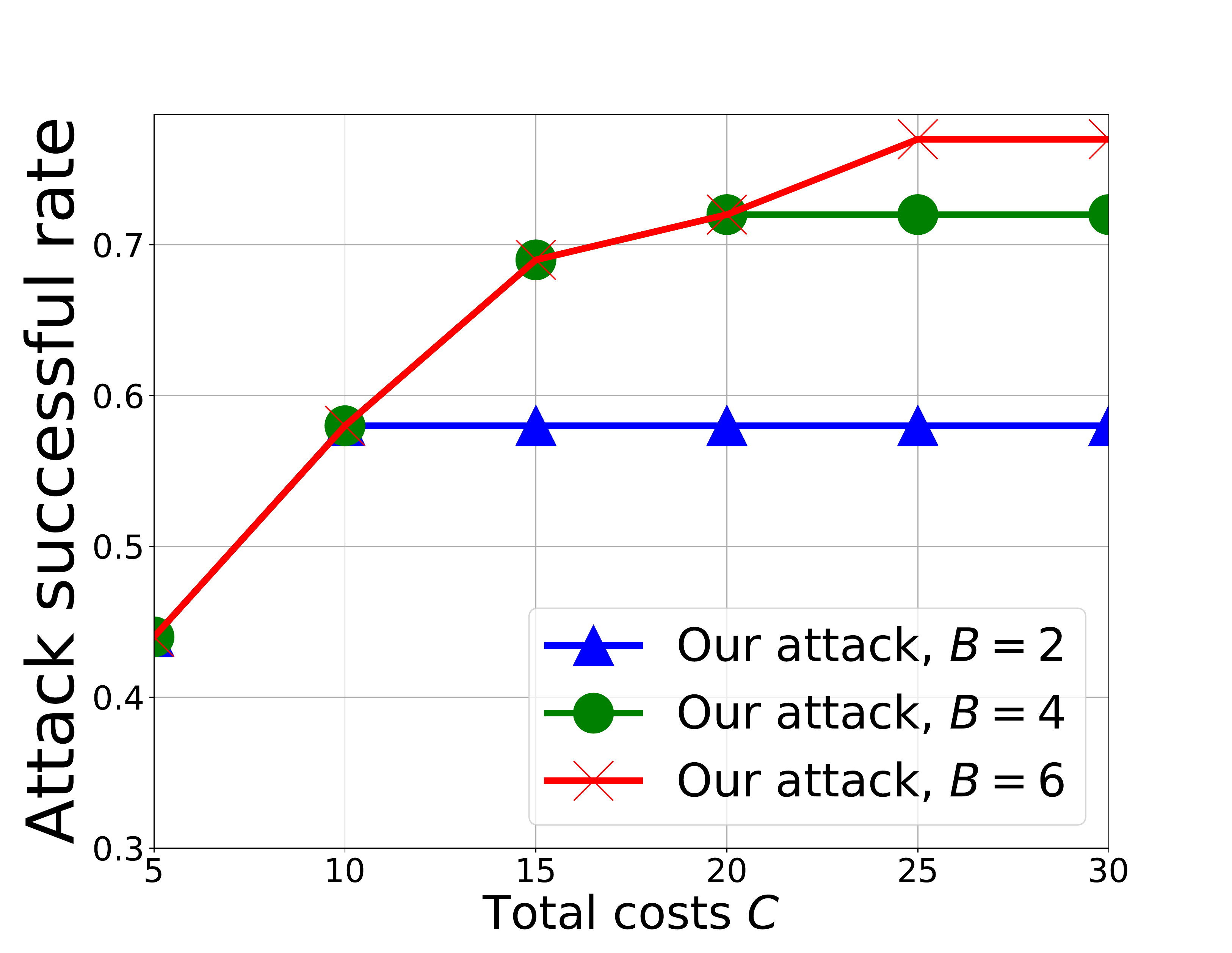}
\end{minipage}%
}%
\subfigure[Citeseer.]{
\begin{minipage}[t]{0.29\linewidth}
\centering
\includegraphics[width=\columnwidth]{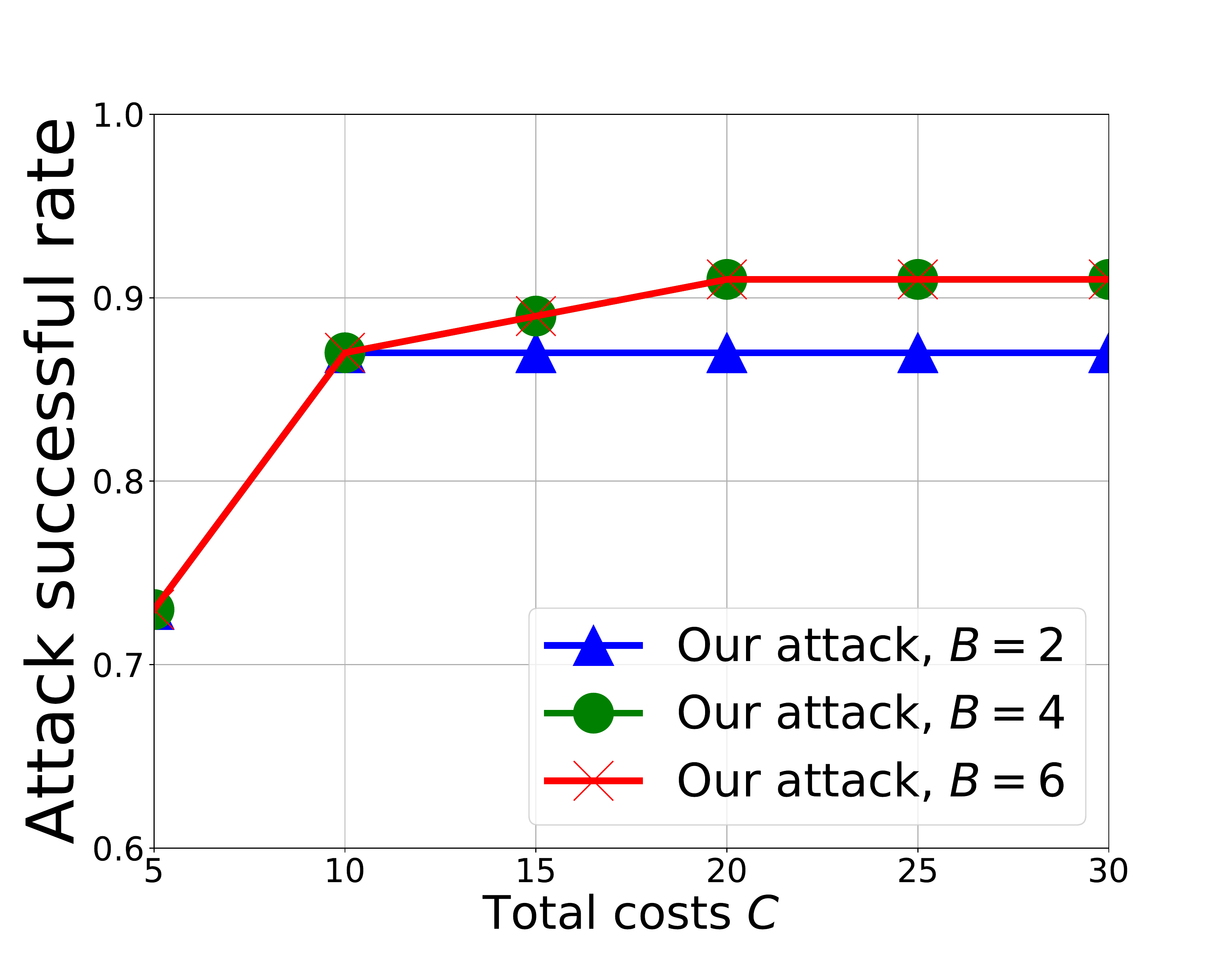}
\end{minipage}%
}%
\subfigure[Pubmed.]{
\begin{minipage}[t]{0.29\linewidth}
\centering
\includegraphics[width=\columnwidth]{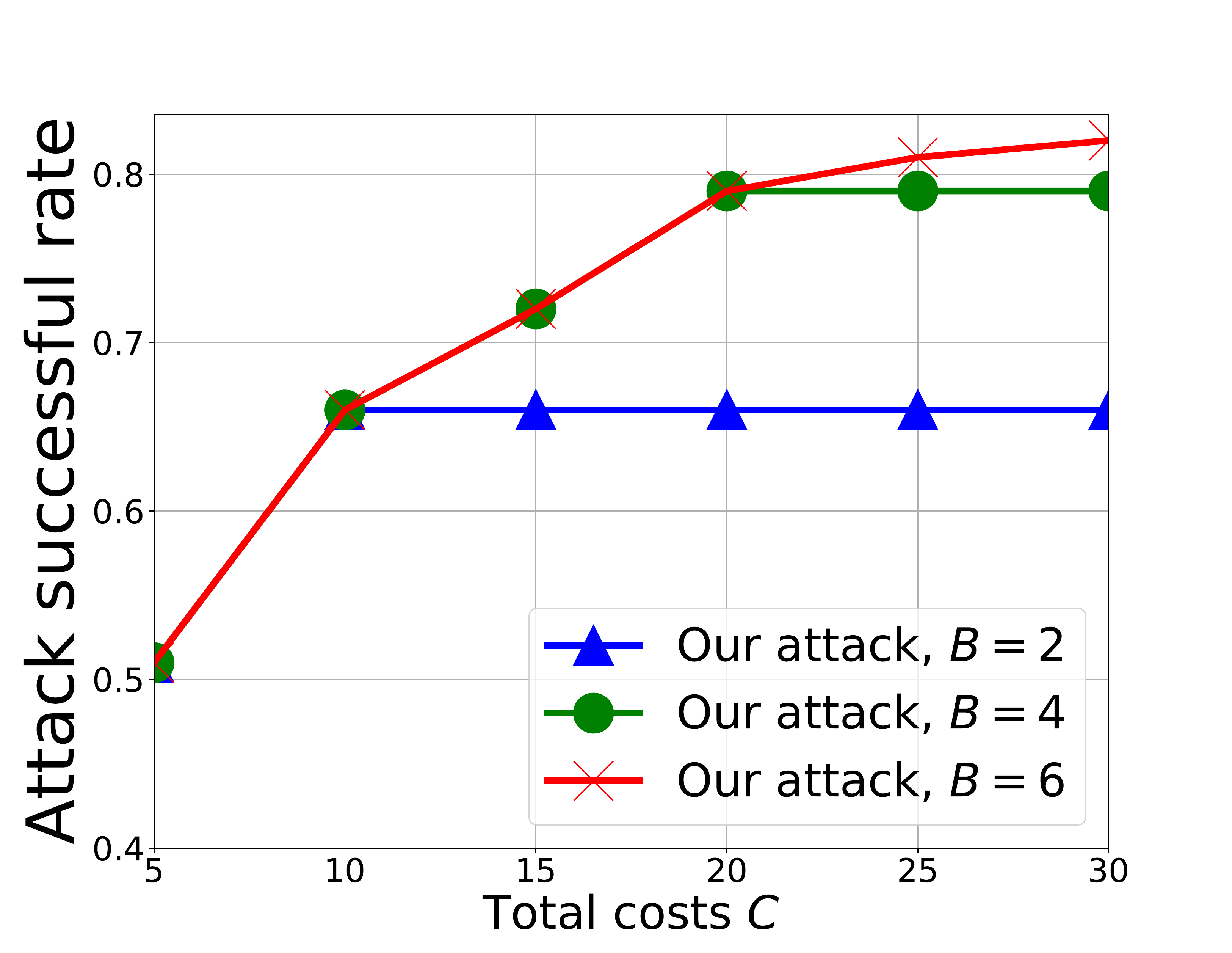}
\end{minipage}%
}%
\centering
\caption{Attack successful rate vs. \#total costs $C$ on GCN for node classification.} 
\label{fig:impact_cost_GCN}
\vspace{-2mm}
\end{figure*}

\vspace{-3mm}
\paragraph{Impact of the total costs.} We study the impact of the total costs $C$ and fix the number of queries $T$ to be $50$. As our attack outperforms the compared attacks, we only study our attack on attacking GCN for simplicity. Note that we have similar observations on attacking SGC and GIN. 

The results are shown in Figure~\ref{fig:impact_cost_GCN}. We observe that the attack successful rate becomes higher when the cost budget is larger. Moreover, note that the cost for perturbing each edge is within [1, 5]. If the cost budget is sufficient, i.e., $C \geq 10, 20, 30$ for $B=2,4,6$, then the attack successful rate is stable, as our attack finds the same space for structure perturbation. In contrast, if the cost budget is insufficient, even the allowed number of perturbed edges is larger, our attack performance cannot be improved, e.g., $C=20$ for $B=4,6$ on Citeseer. The result demonstrates that cost budget is a key factor to affect our attack performance.

\section{Conclusion and Future work}
In this paper, we study black-box attacks to GNNs via manipulating the graph structure. We first formulate our attack as a binary optimization problem, which is NP-hard. Then, we relax and reformulate our attack problem as a bandit optimization problem, and propose a bandit-based attack algorithm and rigorously prove that our attack yields a sublinear regret bound $\mathcal{O}(\sqrt{N}T^{3/4})$ within $T$ queries for attacking a graph with $N$ nodes. Finally, we evaluate our attack against GNN models for both node classification and graph classification. Our results demonstrate both the effectiveness and efficiency of our attack and that our attack significantly outperforms the state-of-the-arts. 

\noindent{\bf Acknowledgments.} 
This work of Wang is supported by the startup funding. The work of Li is partially supported by the National Natural Science Foundation of China (NSFC) under Grant No. 62102028, and China Postdoctoral Science Foundation under Grant No. 2021M700434, and Zhou is supported by the NSFC under Grant No. 61972448.

{\small
\bibliographystyle{ieee_fullname}
\bibliography{ref}
}

\onecolumn
\appendix
\begin{center}
\section*{Appendix}
\end{center}

\section{Notations}
In Table \ref{tab:not}, we summarize the mathematical notations used in this paper .
\begin{table}[ht]
\centering
\caption{Summary of Notations}\label{tab:not}
\begin{tabular}{|c|l|}
 \hline
 Notations & Meanings\\
 \hline
 $G=(\mathcal{V}, \mathcal{E}, \bm{X})$ & The graph $\mathcal{G}$ with node set $\mathcal{V}$, edge set $\mathcal{E}$ and node feature matrix $\bm{X}$\\
 \hline
 $N$ & The size of node set\\
 \hline
 $M$ & The size of node set\\
 \hline
 $\bm{a}_v$ & The adjacency vector of node $v$\\
 \hline
 $\bm{x}_u$ & The feature vector of node $u$\\
 \hline
 $\mathcal{Y}$ & The label set\\
 \hline
 $y_u$ & The label of node $u$\\
 \hline
 $L_C$ & The number of labels\\
 \hline
 $\mathcal{V}_L$ & The training data set\\
 \hline
 $\textrm{AGG}(\cdot)$ & The neighborhood aggregation function of the GNN\\
 \hline
 $\textrm{UPDATE}(\cdot)$ & The node representation update function of the GNN\\
 \hline
 $\bm{l}_v^{(k)}$ & The aggregated representation vector of node $v$ in layer $k$\\
 \hline
$\bm{h}_v^{(k)}$ & $v$'s representation  in the $k$-th layer \\
 \hline
 $\Theta$ & The GNN model parameters\\
 \hline
 $f_{\Theta}(\bm{a}_v)$ & $v$'s model output\\
 \hline
 $B$ & The maximum number of perturbed edges\\
 \hline
 $C$ & Cost budget\\
 \hline
 $\bm{s}_v$ & The adversarial structure perturbation vector for the targeted node $v$\\
 \hline
 $\bm{\hat{s}}_v$ & The relaxed vector of $\bm{s}_v$\\
 \hline
 $\bm{c}_v$ & The cost vector associated with $v$ on edges\\
 \hline
 $L(\bm{a}_v)$ & The loss function for the targeted node $v$ without attack\\
 \hline
 $L(\bm{a}_v \oplus \bm{s}_v)$ & The attack loss for the targeted node $v$ under perturbation $\bm{s}_v$\\ 
 \hline
 $C_L$ & The Lipschitz constant with respect to loss $L(\cdot)$\\
 \hline
 $\mathtt{Reg}(T)$ & The incurred regret within $T$ rounds\\
 \hline
 $\mathcal{W}$ & The arm set consisting of feasible $\bm{\hat{s}}_v$\\
 \hline
 $\mathcal{S}$ & The unit sphere with respect to $2$-norm\\
 \hline
 $\mathbb{B}$ & The unit ball with respect to $2$-norm\\
 \hline
 $\tilde{L}(\cdot)$ & The smooth loss function \\
 \hline
 $\bm{v}_t$ & The prior vector at round $t$\\
 \hline
 $\bm{u}_t$ & The random unit vector\\
 \hline
 $\bm{\hat{g}}$ & The estimated gradient using in PGD\\
 \hline
\end{tabular}
\end{table}

\newpage
\section{Proof of Lemma \ref{le:smoothL}}
\begin{proof}
We first consider the case where $N = 1$, i.e., scalar setting. Thus, we have $|u| = 1$ and $u$ takes value in $\{-1,1\}$. When uniformly sampling its value, we have,
\begin{equation}
\begin{split}
\nabla\hat{L}(\bm{\hat{s}}_v) &= \mathbb{E}_{u\in\{-1,1\}}[\frac{1}{\delta}L(\bm{\hat{s}}_v+\delta u)u]\\
& = \frac{1}{2}\frac{L(s+\delta)}{\delta} - \frac{1}{2}\frac{L(s-\delta)}{\delta} = L'(s).
\end{split}
\end{equation}
For $N>1$ setting, the lemma can be proved using Stoke's theorem \cite{stewart2012essential}. For more details, please refer to \cite{flaxman2004online}.
\end{proof}

\section{Proof of Lemma \ref{le:relaxreg}}
\begin{proof}
We start the proof from Lipschitz continuity assumed for $L(\cdot)$. Based on Eq. (\ref{eq:Lip}), we have
\begin{equation*}
\begin{split}
\mathbb{E}[|L(\bm{s}_v) - L(\bm{\hat{s}}_v^t)|] &\stackrel{(a)}{\le} \mathbb{E}[C_L ||\bm{s}_v - \bm{\hat{s}}_v^t||_2]\\
&\stackrel{(b)}{=} C_L \mathbb{E}[\sqrt{\sum_{i=1}^{N}(\bm{s}_v(i) - \bm{\hat{s}}_v^t(i))^2}]\\
&\stackrel{(c)}{\le} C_L \sqrt{\sum_{i=1}^{N}\mathbb{E}[(\bm{s}_v(i) - \bm{\hat{s}}_v^t(i))^2]}\\
&\stackrel{(d)}{=} C_L \sqrt{\sum_{i=1}^{N}\mathbb{E}[(\bm{s}_v(i) - \mathbb{E}[\bm{s}_v(i)])^2]}\\
&\stackrel{(e)}{=} C_L \sqrt{\sum_{i=1}^{N}\mathbb{E}[\bm{s}_v(i)^2] - \mathbb{E}[\bm{s}_v(i)]^2}\\
&\stackrel{(g)}{\le} C_L \sqrt{\sum_{i=1}^{N}\mathbb{E}[\bm{s}_v(i)^2]}
\stackrel{(f)}{=} C_L \sqrt{||\bm{\hat{s}}_v^t||_1}\\
&\stackrel{(h)}{\le} C_L\sqrt{\sqrt{N}||\bm{\hat{s}}_v^t||_2}
\stackrel{(i)}{\le} C_L\sqrt{\sqrt{N}||\bm{v}_{t-1} - \eta \bm{\hat{g}}_{t-1}||_2} \\
&\stackrel{(j)}{\le} C_L\sqrt{\sqrt{N}||\bm{v}_{t-1}||_2 + \eta\sqrt{N}||\bm{\hat{g}}_{t-1}||_2}\\
&\stackrel{(k)}{\le} C_L{N}^{3/4}\sqrt{1+\frac{\eta}{\delta}},
\end{split}
\end{equation*}
where (a) is due to Lipschitz continuity of attack loss function $L(\cdot)$, (b) is due to the definition of $2$-norm, (c) holds due to applying Jensen's inequality, (d) is due to the random rounding between $\bm{\hat{s}}_v^t$ and $\bm{s}_v$, (e) is due to definition of variance related to random variable $\bm{s}_v(i)$, i.e., $\mathbb{E}[(\bm{s}_v(i) - \mathbb{E}[\bm{s}_v(i)])^2] = \mathbf{Var}[\bm{s}_v(i)] = \mathbb{E}[\bm{s}_v(i)^2] - \mathbb{E}[\bm{s}_v(i)]^2$. In (g), we drop the negative term that does not affect the bound. In (f), we use the fact that each component $\bm{s}_v(i)\in\{0,1\}$ is bounded by one such that $\bm{s}_v(i)^2 = \bm{s}_v(i)$. Besides, we apply the definition of $1$-norm. In (h), we use the inequality $||\bm{\hat{s}}_v^t||_1 \le \sqrt{N}||\bm{\hat{s}}_v^t||_2$ for $1$-norm and $2$-norm. In (i), we substitute the $\bm{\hat{s}}_v^t$ by the projected gradient descent equation and apply the non-extensive property of the projection onto convex set. In (j), we apply the triangle inequality. In (k), we expand the $2$-norm using the definition of $\bm{v}_{t-1}$ and $\bm{\hat{g}}_{t-1}$. 
\end{proof}

Similar claim can be found in $\mathtt{bOGD}$ \cite{lesage2020online}. However, the differences of $\mathtt{bOGD}$ \cite{lesage2020online} and our result in Lemma \ref{le:relaxreg} are two-fold:
\begin{inparaenum}[1)]
\item we consider the bandit setting of OCO, which requires to estimate gradient while $\mathtt{bOGD}$ performs based on the derived gradient when observing the loss function;
\item we only assume that the loss function is Lipschitz continuous related to $2$-norm while $\mathtt{bOGD}$ additionally assume the loss is Lipschitz continuous related to $1$-norm and $2$-norm.
\end{inparaenum}

\section{Proof of Theorem \ref{th:regbound}}
\begin{proof}
Suppose the arm set $\mathcal{W}$ satisfies $R_1\mathbb{B}\subseteq\mathcal{W}\subseteq R_2\mathbb{B}$. We begin the proof from the regret definition. 
\begin{equation}\label{eq:regder}
\begin{split}
& \mathtt{Reg}(T) = \mathbb{E}[\sum_{t=1}^TL(\bm{s}_v)] - TL(\bm{s}_v^*)\\
&\stackrel{(a)}{\le} \mathbb{E}[\sum_{t=1}^TL(\bm{s}_v)] - TL(\bm{\hat{s}}_v^*)\\
&\stackrel{(b)}{\le} \mathbb{E}[\sum_{t=1}^TL(\bm{s}_v)] - \min_{\bm{w}\in(1-\alpha)\mathcal{W}}\sum_{t=1}^TL(\bm{w}) + \min_{\bm{w}\in(1-\alpha)\mathcal{W}}\sum_{t=1}^TL(\bm{w}) - TL(\bm{\hat{s}}_v^*)\\
&\stackrel{(c)}{\le} 
\mathbb{E}[\sum_{t=1}^TL(\bm{s}_v)] - \min_{\bm{w}\in(1-\alpha)\mathcal{W}}\sum_{t=1}^TL(\bm{w}) + 2\alpha T\\
&\stackrel{(d)}{\le}
 \mathbb{E}\Big[\sum_{t=1}^T[L(\bm{s}_v) - L(\bm{\hat{s}}_v^t)]\Big] + \mathbb{E}[\sum_{t=1}^TL(\bm{\hat{s}}_v^t)]  - \min_{\bm{w}\in(1-\alpha)\mathcal{W}}\sum_{t=1}^TL(\bm{w}) + 2\alpha T\\
&\stackrel{(e)}{\le}
\mathbb{E}[\sum_{t=1}^TL(\bm{\hat{s}}_v^t)] - \min_{\bm{w}\in(1-\alpha)\mathcal{W}}\sum_{t=1}^TL(\bm{w}) + 2\alpha T + C_LT{N}^{3/4}\sqrt{1+\frac{\eta}{\delta}}\\
&\stackrel{(f)}{\le}
\mathbb{E}[\sum_{t=1}^T\tilde{L}(\bm{v}^t)] - \min_{\bm{w}\in(1-\alpha)\mathcal{W}}\sum_{t=1}^T\tilde{L}(\bm{w}) + \mathbb{E}\Big[\sum_{t=1}^T[L(\bm{\hat{s}}_v^t) - \tilde{L}(\bm{v}^t)]\Big] \\
& + \min_{\bm{w}\in(1-\alpha)\mathcal{W}}\sum_{t=1}^T[\tilde{L}(\bm{w})- L(\bm{w})] + 2\alpha T + C_LT{N}^{3/4}\sqrt{1+\frac{\eta}{\delta}}.
\end{split}
\end{equation}

In (a), we consider that $L(\bm{s}_v^*)\ge L(\bm{\hat{s}}_v^*)$ holds for the relaxed variable $\bm{\hat{s}}_v^*$. Generally speaking, the attack loss achieved by the integer optimal solution is greater than or equal to the one of the relaxed continuous optimal solution. In (b), we consider the $(1-\alpha)$-projection in our algorithm (i.e., line 7). In (c), we apply with the result of Lemma \ref{le:scalepro}. In (d), we consider the relaxed variable $\bm{\hat{s}}_v^t$. In (e), we apply with the result of Lemma \ref{le:relaxreg}. In (f), we adapt the terms to make convenience of using the result of Lemma \ref{le:sgdregret}.

We define a smooth function $\tilde{L}(\cdot)$ to approximate the original loss function $L(\cdot)$ as follows,
\begin{equation}
\tilde{L}(\bm{v}) = \mathbb{E}_{\bm{u}\in\mathbb{B}}[L(\bm{v}+\delta\bm{u})].
\end{equation}
When $\delta$ is small, $\tilde{L}(\bm{v})\approx L(\bm{v})$. Thus, $\tilde{L}(\bm{v})$ and $L(\bm{v})$ share similar gradient at any $\bm{v}$. According to the Lipschitz continuity in Eq. (\ref{eq:Lip}), we have,
\begin{equation}
|\tilde{L}(\bm{w})- L(\bm{w})| \le C_L\delta,
\end{equation}
for all $\bm{w}\in\mathcal{W}$. Due to $\bm{\hat{s}}_v^t = \bm{v}^t + \delta \bm{u}^t$, we obtain,
\begin{equation}
|L(\bm{\hat{s}}_v^t) - L(\bm{v}^t)| = |L(\bm{v}^t + \delta \bm{u}^t) - L(\bm{v}^t)| \le C_L\delta,
\end{equation}
where we further have by applying triangle inequality,
\begin{equation}
|L(\bm{\hat{s}}_v^t) - \tilde{L}(\bm{v}^t)| \le |L(\bm{\hat{s}}_v^t) - L(\bm{v}^t)| + |L(\bm{v}^t) - \tilde{L}(\bm{v}^t)| \le 2C_L\delta,
\end{equation}
for all $t\in[T]$. To continue bounding $\mathtt{Reg}(T)$ in Eq. (\ref{eq:regder}).(f), we have,
{\footnotesize
\begin{equation}
\begin{split}
& \mathtt{Reg}(T) \le \sum_{t=1}^T\tilde{L}(\bm{v}^t) - \min_{\bm{w}\in(1-\alpha)\mathcal{W}}\sum_{t=1}^T\tilde{L}(\bm{w}) + 3C_L\delta T + 2\alpha T \\
& \qquad \quad +  C_LT{N}^{3/4}\sqrt{1+\frac{\eta}{\delta}}.
\end{split}
\end{equation}
}
At last, we bound the regret $\sum_{t=1}^T\tilde{L}(\bm{v}^t) - \min_{\bm{w}\in(1-\alpha)\mathcal{W}}\sum_{t=1}^T\tilde{L}(\bm{w})$, which corresponds to the stochastic gradient descent in line 7 of Algorithm \ref{alg:bb_ocob} where the gradient is $\frac{N}{\delta}L(\bm{\hat{s}}_v^t)\bm{u}^t$ and the feasible domain is $(1-\alpha)\mathcal{W}$. Note that we have $(1-\alpha)\mathcal{W}\subseteq\mathcal{W}\subseteq R_2\mathbb{B}$ for any $0<\alpha<1$. It is easy to see that $||\frac{N}{\delta}L(\bm{\hat{s}}_v^t)\bm{u}^t||_2\le\frac{N}{\delta}$. According to Lemma \ref{le:sgdregret}, we have
\begin{equation}
\mathbb{E}[\sum_{t=1}^T\tilde{L}(\bm{v}^t)] - \min_{\bm{w}\in(1-\alpha)\mathcal{W}}\sum_{t=1}^T\tilde{L}(\bm{w}) \le \frac{N}{\delta}R_2\sqrt{T}.
\end{equation}
In summary, we can bound $\mathtt{Reg}(T)$ as,
{\footnotesize
\begin{equation}
\begin{split}
\mathtt{Reg}(T) &\le \frac{N}{\delta}R_2\sqrt{T} + 3C_L\delta T + 2\alpha T
+ C_LT{N}^{3/4}\sqrt{1+\frac{\eta}{\delta}}\\
&\stackrel{(a)}{\le}\frac{N}{\delta}R_2\sqrt{T} + 3C_L\delta T + 2\alpha T 
+ C_LT{N}^{3/4}\sqrt{\frac{R_2}{{N}\sqrt{T}}}\\
&\stackrel{(b)}{\le}
\frac{N}{\delta}R_2\sqrt{T} + 3C_L\delta T + 2\alpha T 
+ C_L\sqrt{R_2}{N}^{1/2}T^{3/4}\\
&\stackrel{(c)}{\le}
\frac{N}{\delta}R_2\sqrt{T} + \frac{3(C_LR_1+1)}{R_1}\delta T
 + C_L\sqrt{R_2}{N}^{1/2}T^{3/4}\\
&\stackrel{(d)}{\le}
2\sqrt{\frac{3R_2(C_LR_1+1){N}}{R_1}}T^{3/4} 
+ C_L\sqrt{R_2}{N}^{1/2}T^{3/4}\\
&\stackrel{(e)}{=}\Lambda \sqrt{N}T^{3/4} = \mathcal{O}(\sqrt{N}T^{3/4}).
\end{split}
\end{equation}
}
In (a), we set the step $\eta = \frac{R_2}{\sqrt{T}{N}/\delta} - \delta$. We further simplify the last term, which leads to (b). In (c), we set $\alpha = \frac{\delta}{R_1}$, which can ensure $\bm{\hat{s}}_v^t\in\mathcal{W}$ \cite{flaxman2004online}. In (d), we set parameter $\delta = T^{-1/4}\sqrt{\frac{R_1R_2{N}}{3(C_LR_1+1)}}$ to minimize the r.h.s. of (c). In (e), we define the leading constant as $\Lambda = 2\sqrt{\frac{3R_2(C_LR_1+1)}{R_1}} + C_L\sqrt{R_2}$. Based on above deduction, we can prove this theorem.  
\end{proof}

\begin{figure*}[!t]
\centering
\subfigure[GCN with Cora.]{
\begin{minipage}[t]{0.24\linewidth}
\centering
\includegraphics[width=\columnwidth]{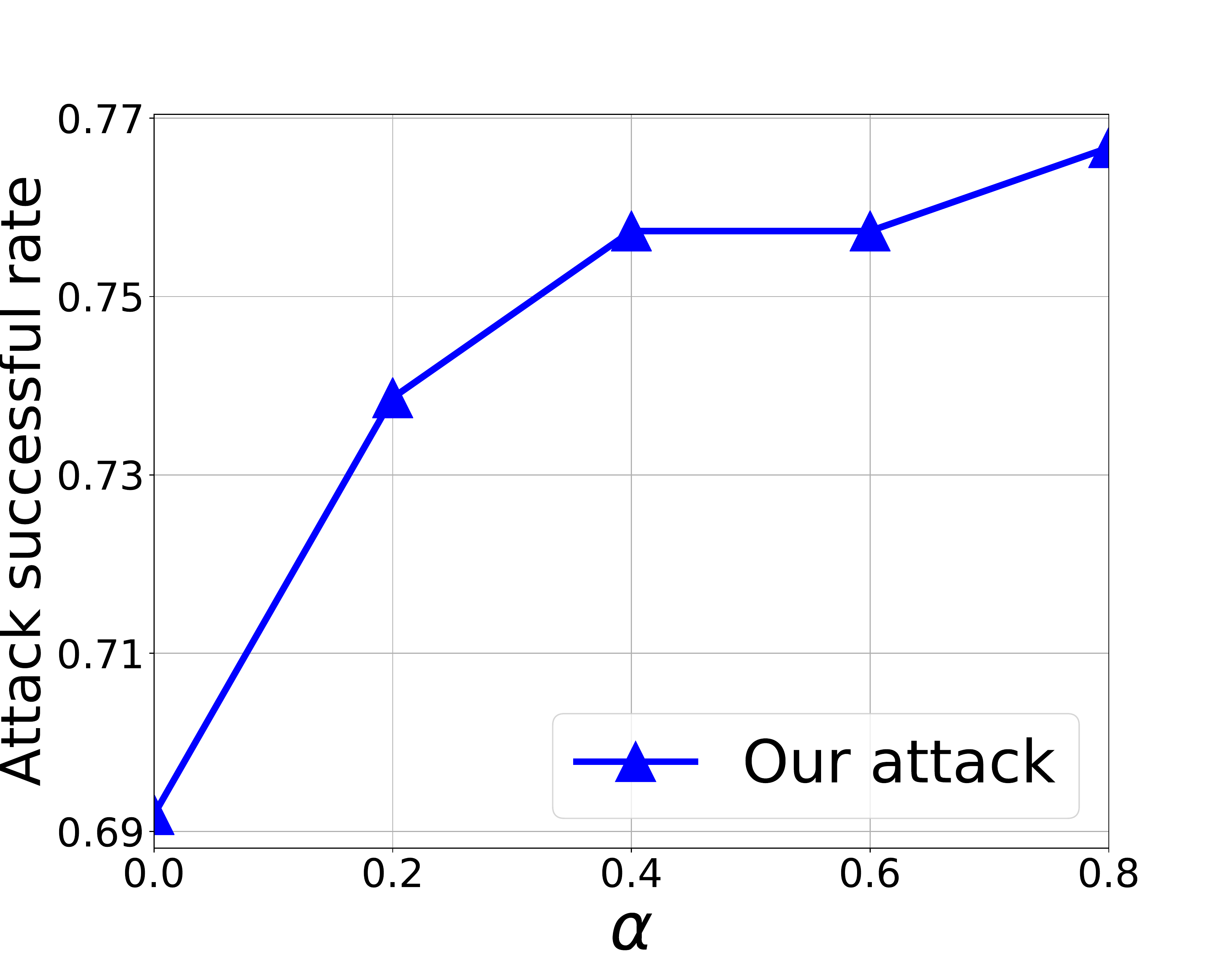}
\end{minipage}%
}%
\subfigure[GCN with Citeseer.]{
\begin{minipage}[t]{0.24\linewidth}
\centering
\includegraphics[width=\columnwidth]{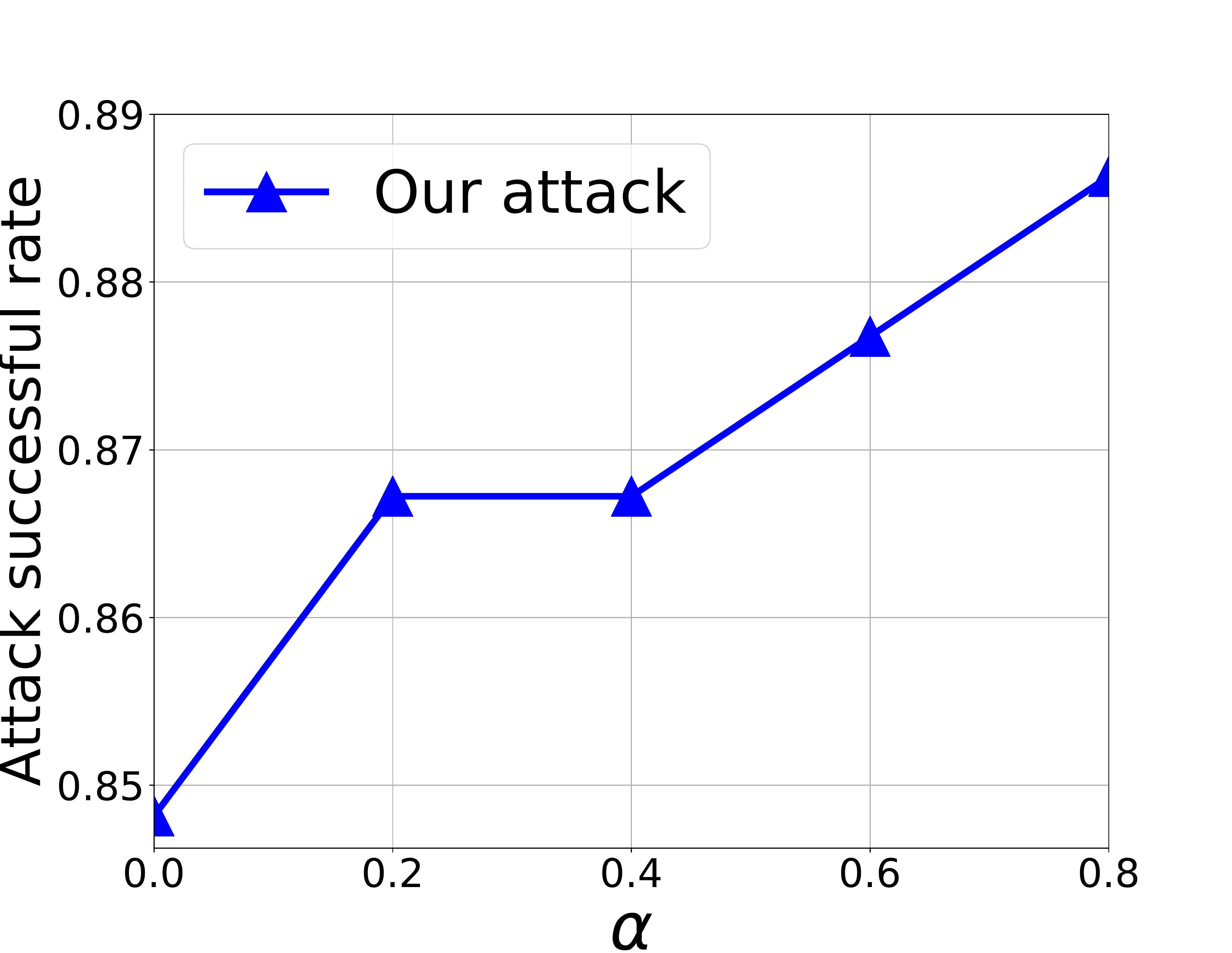}
\end{minipage}%
}%
\subfigure[GCN with Pubmed.]{
\begin{minipage}[t]{0.24\linewidth}
\centering
\includegraphics[width=\columnwidth]{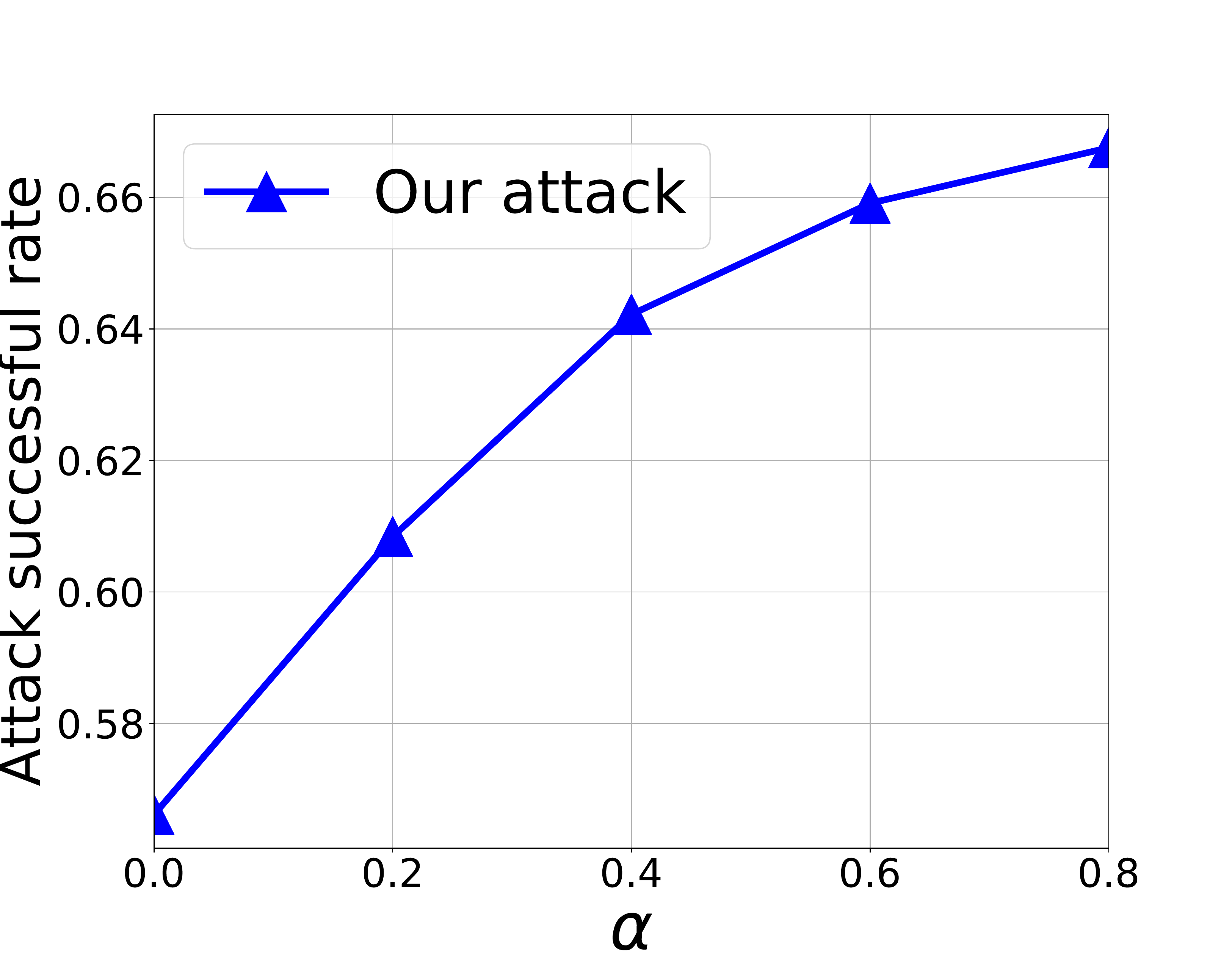}
\end{minipage}%
}%
\subfigure[GIN with MNIST.]{
\begin{minipage}[t]{0.24\linewidth}
\centering
\includegraphics[width=\columnwidth]{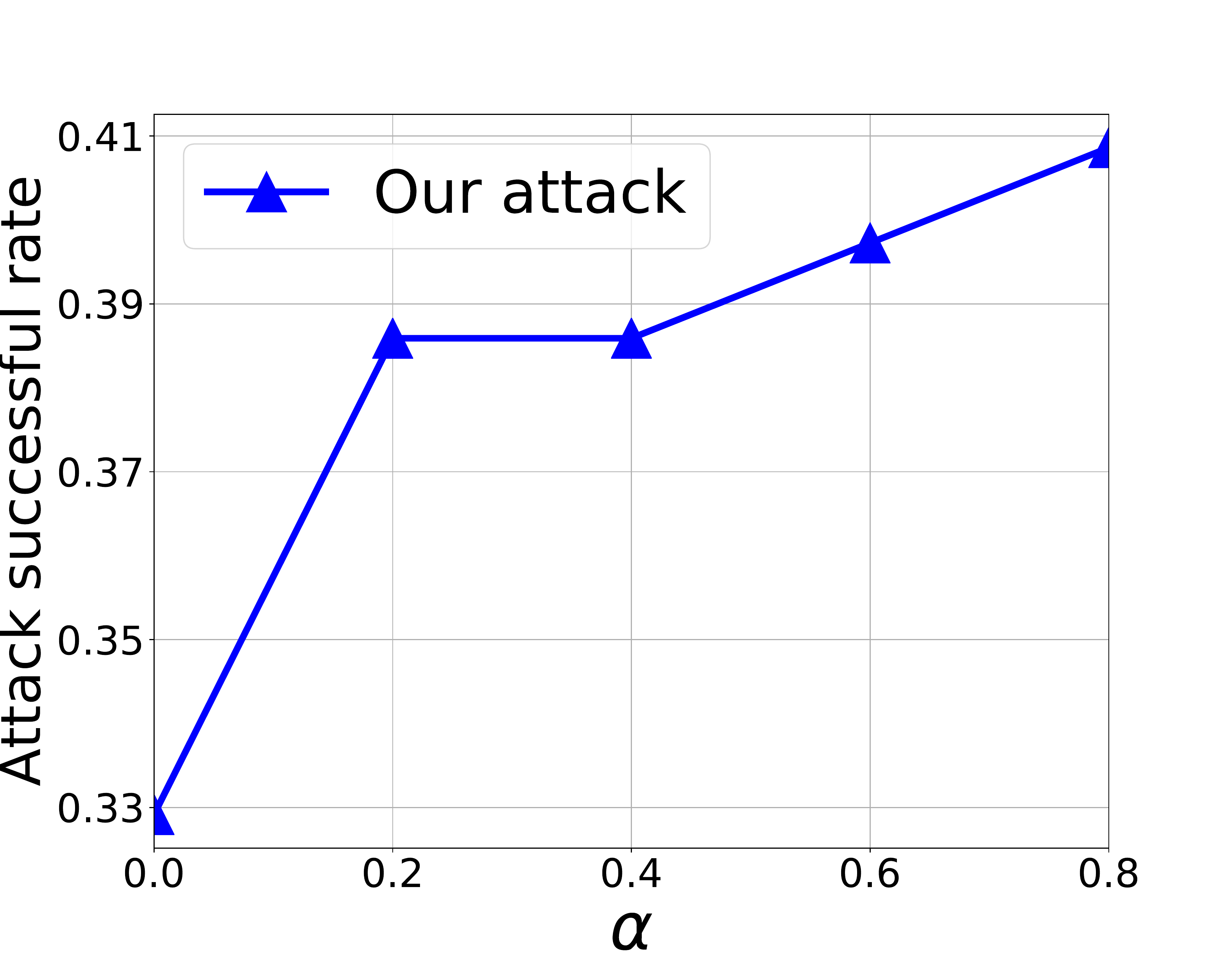}
\end{minipage}%
}%
\\
\subfigure[SGC with Cora.]{
\begin{minipage}[t]{0.24\linewidth}
\centering
\includegraphics[width=\columnwidth]{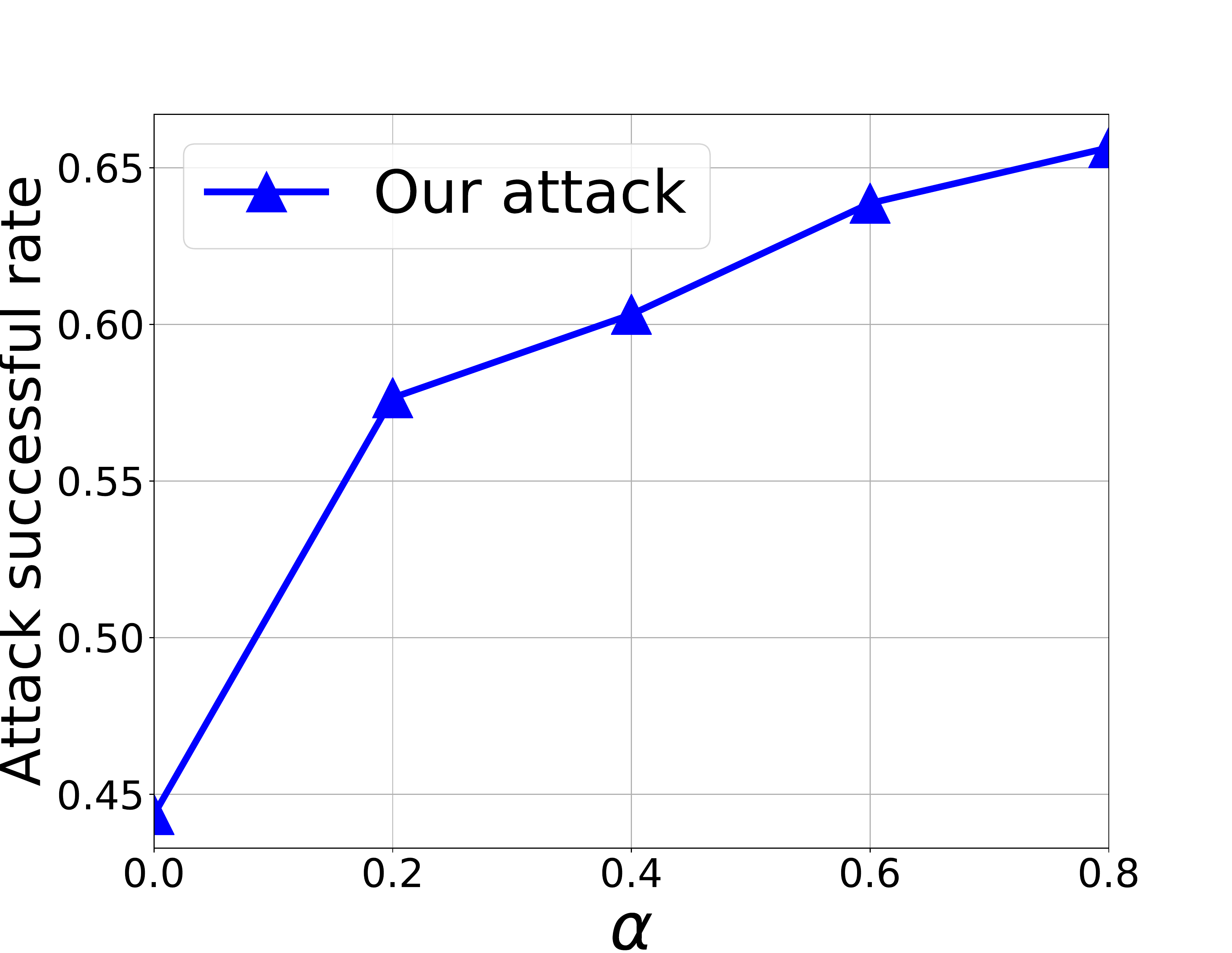}
\end{minipage}%
}%
\subfigure[SGC with Citeseer.]{
\begin{minipage}[t]{0.24\linewidth}
\centering
\includegraphics[width=\columnwidth]{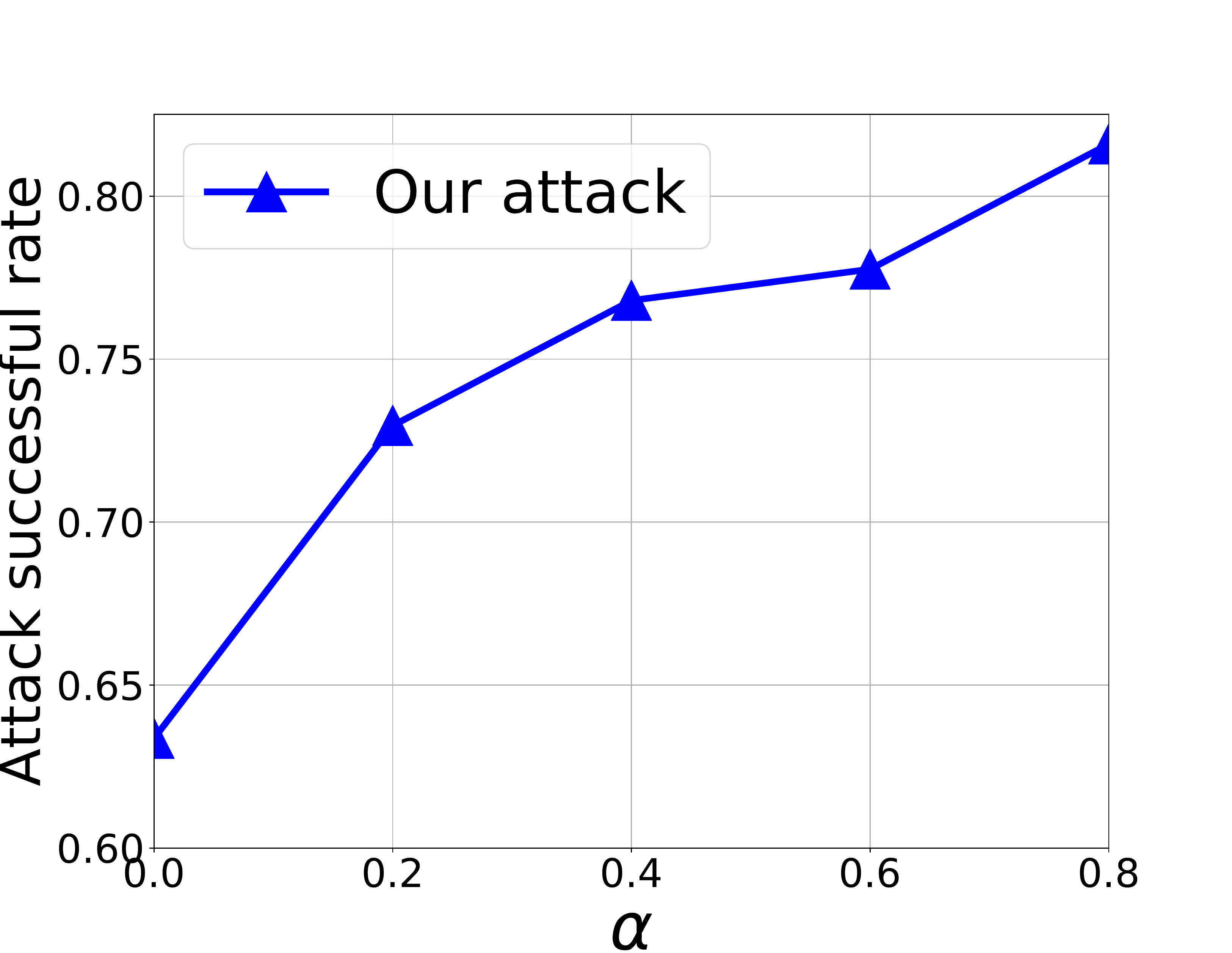}
\end{minipage}%
}%
\subfigure[SGC with Pubmed.]{
\begin{minipage}[t]{0.24\linewidth}
\centering
\includegraphics[width=\columnwidth]{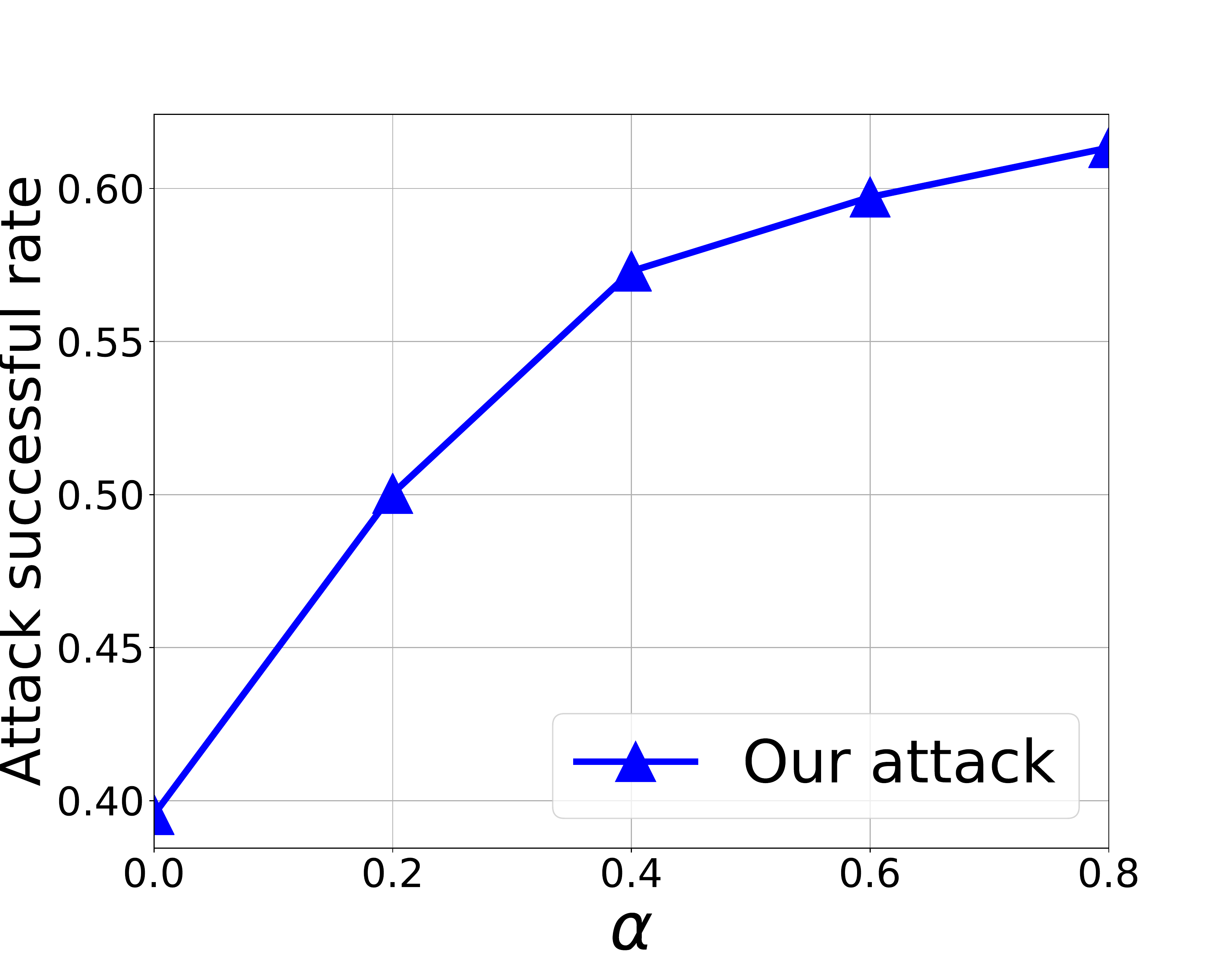}
\end{minipage}%
}%
\subfigure[GIN with CIFAR10.]{
\begin{minipage}[t]{0.24\linewidth}
\centering
\includegraphics[width=\columnwidth]{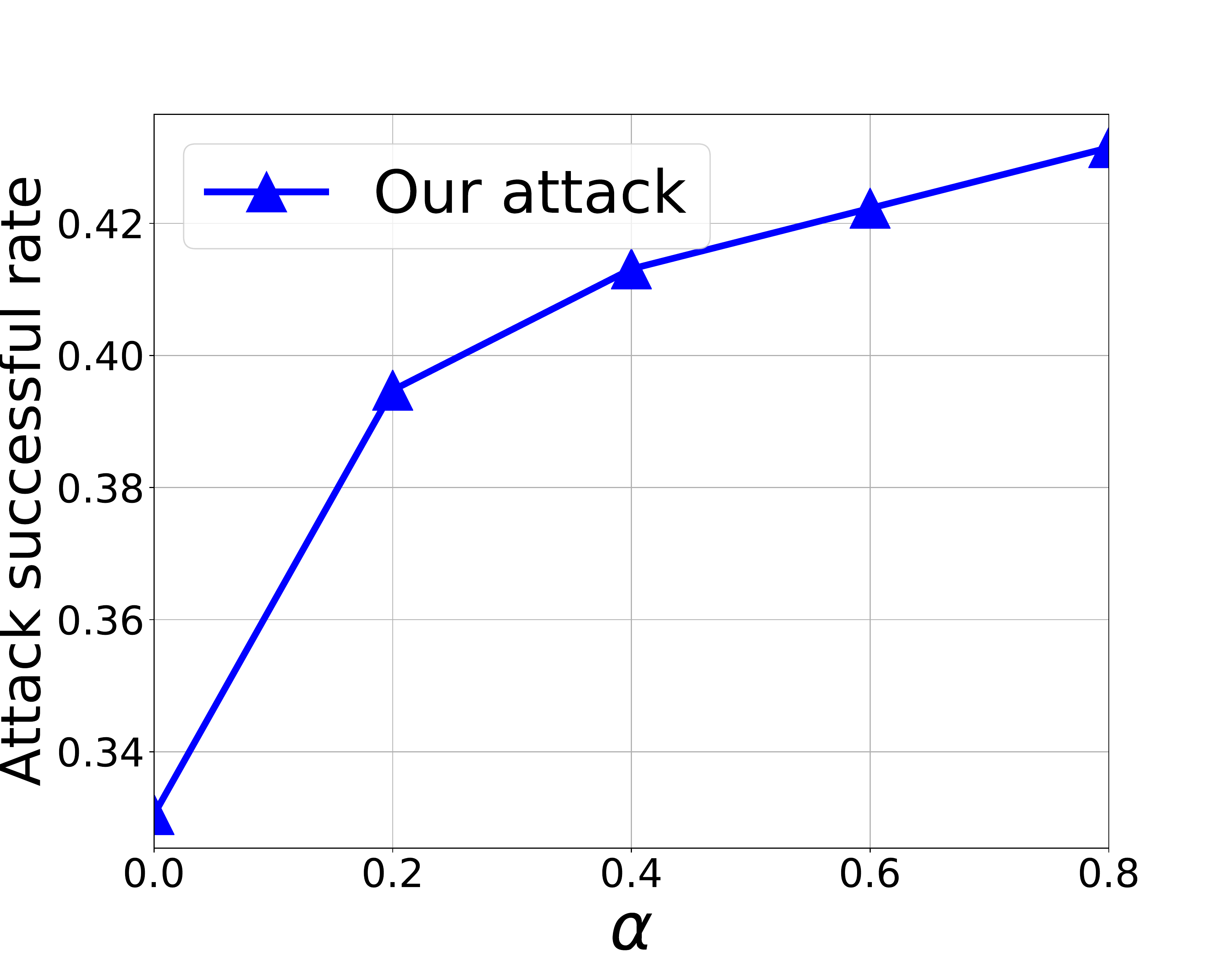}
\end{minipage}%
}%
\centering
\caption{Sensitivity of hyperparameter $\alpha$.} 
\label{fig:alpha}
\end{figure*}

\begin{figure*}[!t]
\centering
\subfigure[GCN with Cora.]{
\begin{minipage}[t]{0.24\linewidth}
\centering
\includegraphics[width=\columnwidth]{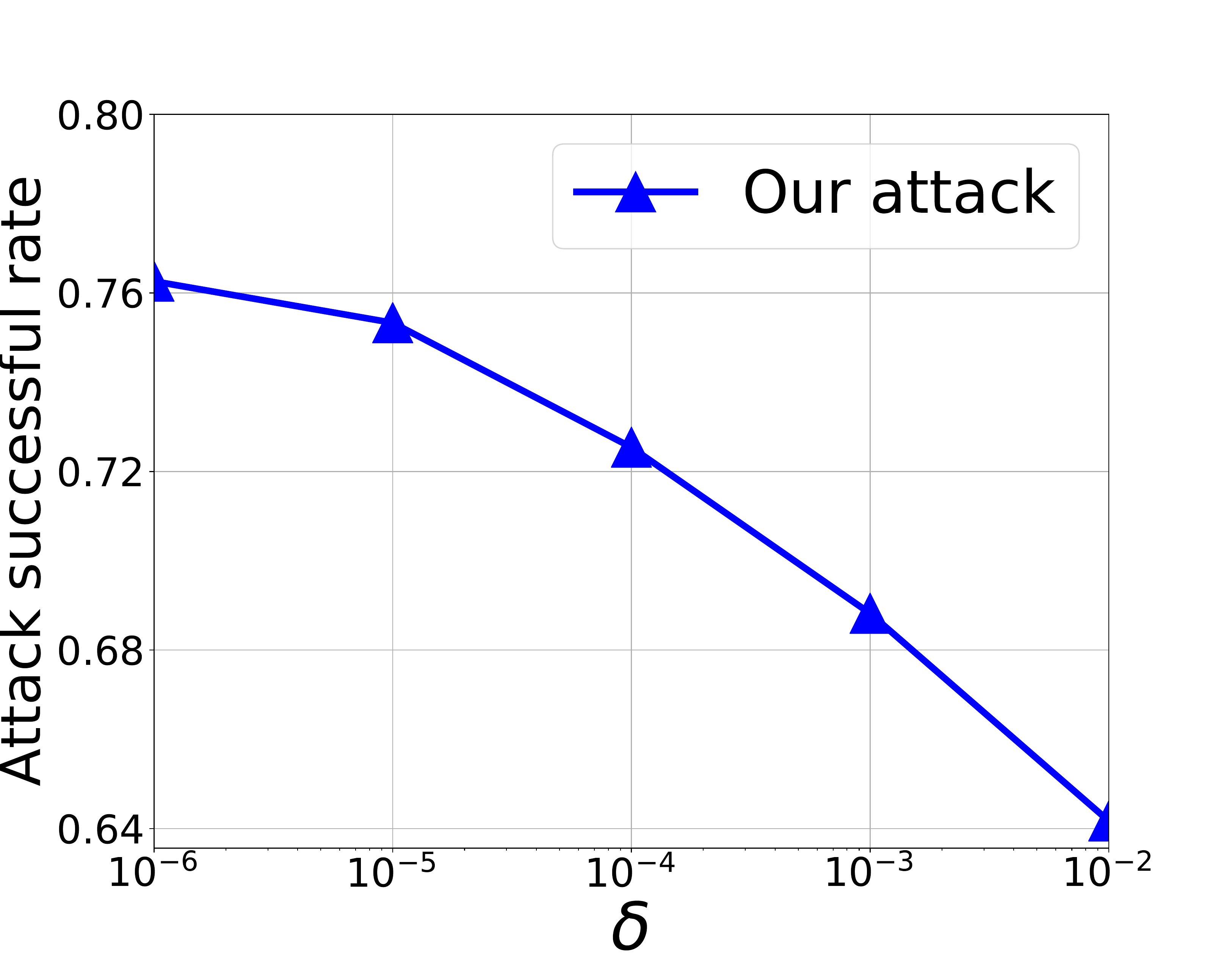}
\end{minipage}%
}%
\subfigure[GCN with Citeseer.]{
\begin{minipage}[t]{0.24\linewidth}
\centering
\includegraphics[width=\columnwidth]{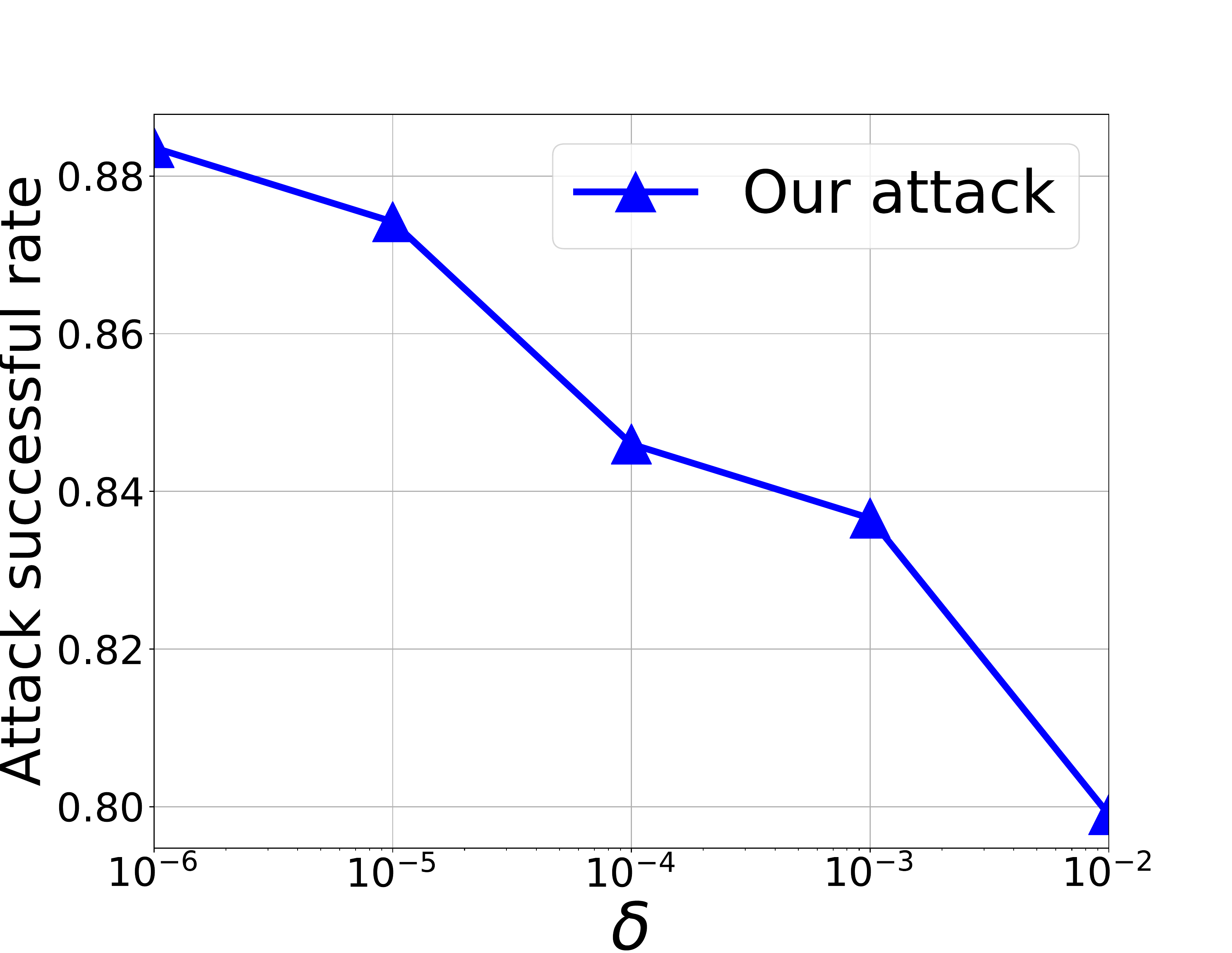}
\end{minipage}%
}%
\subfigure[GCN with Pubmed.]{
\begin{minipage}[t]{0.24\linewidth}
\centering
\includegraphics[width=\columnwidth]{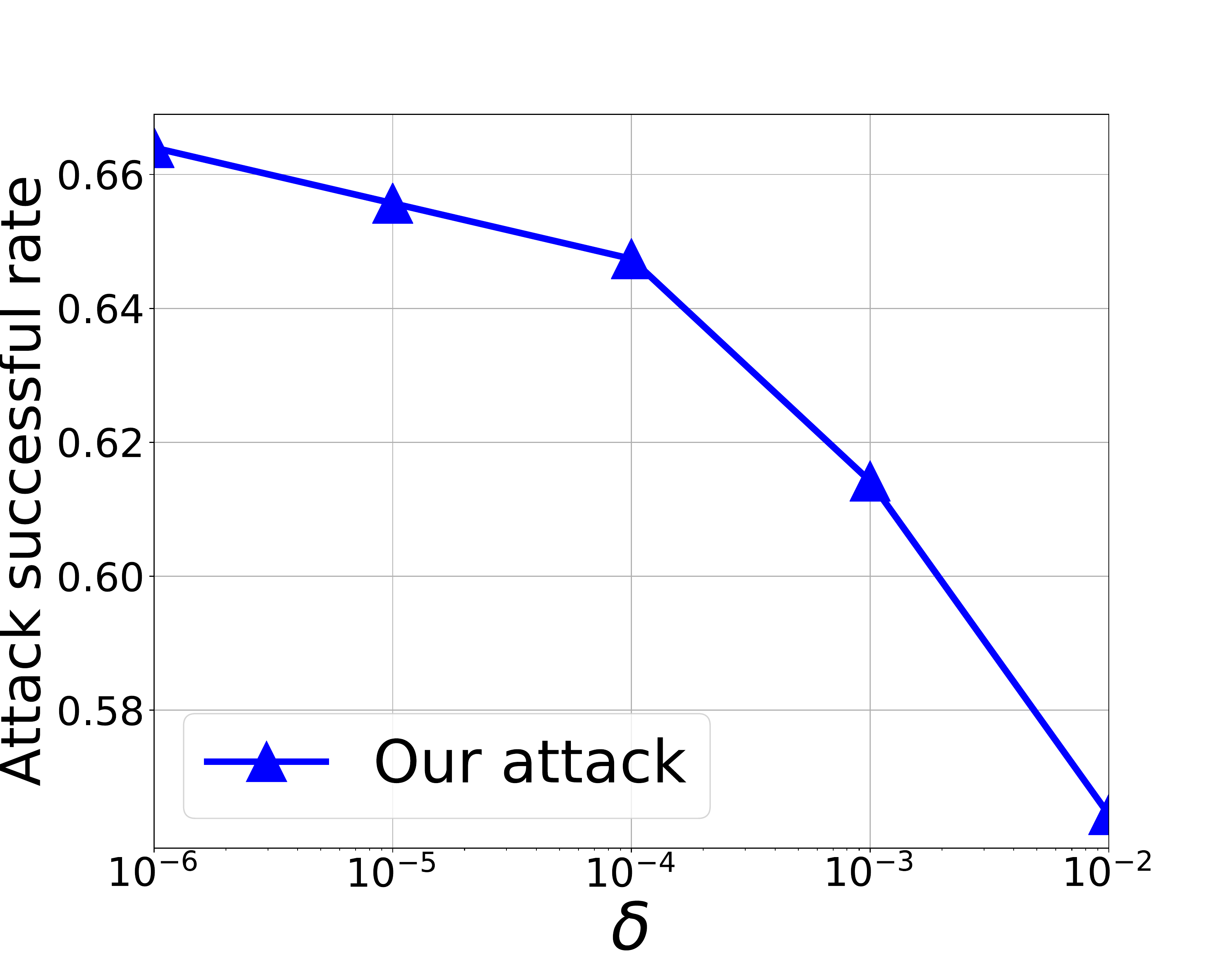}
\end{minipage}%
}%
\subfigure[GIN with MNIST.]{
\begin{minipage}[t]{0.24\linewidth}
\centering
\includegraphics[width=\columnwidth]{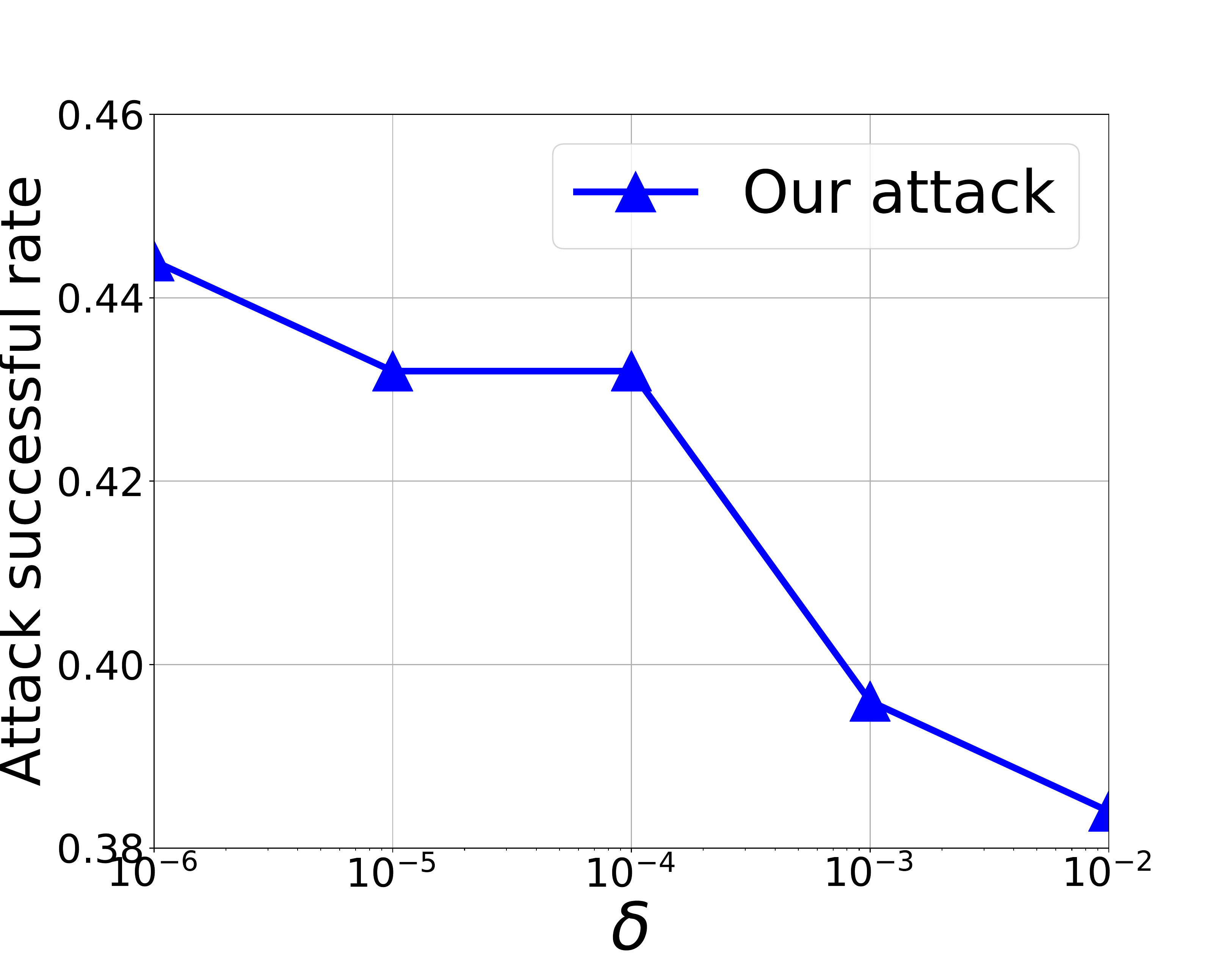}
\end{minipage}%
}%
\\
\subfigure[SGC with Cora.]{
\begin{minipage}[t]{0.24\linewidth}
\centering
\includegraphics[width=\columnwidth]{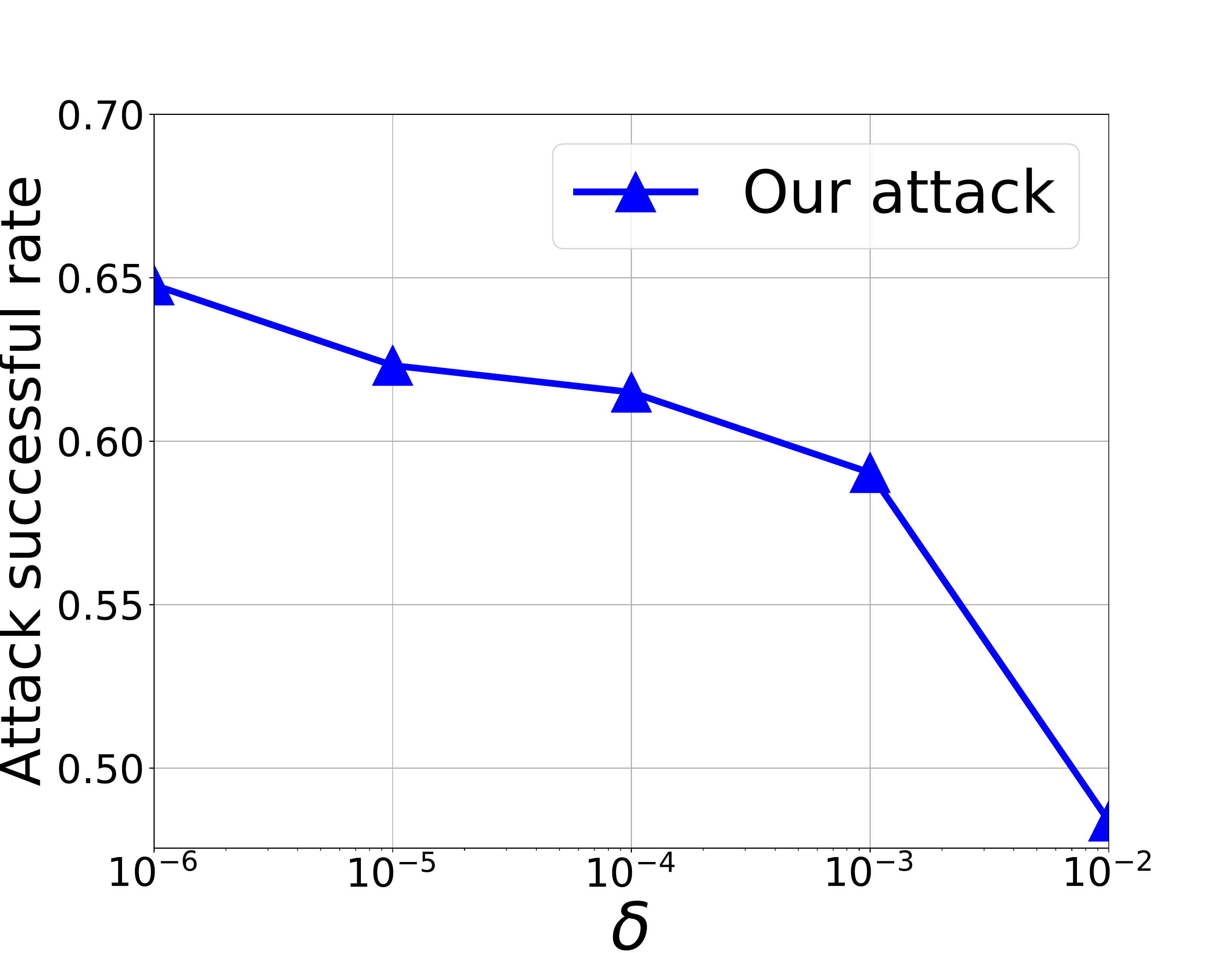}
\end{minipage}%
}%
\subfigure[SGC with Citeseer.]{
\begin{minipage}[t]{0.24\linewidth}
\centering
\includegraphics[width=\columnwidth]{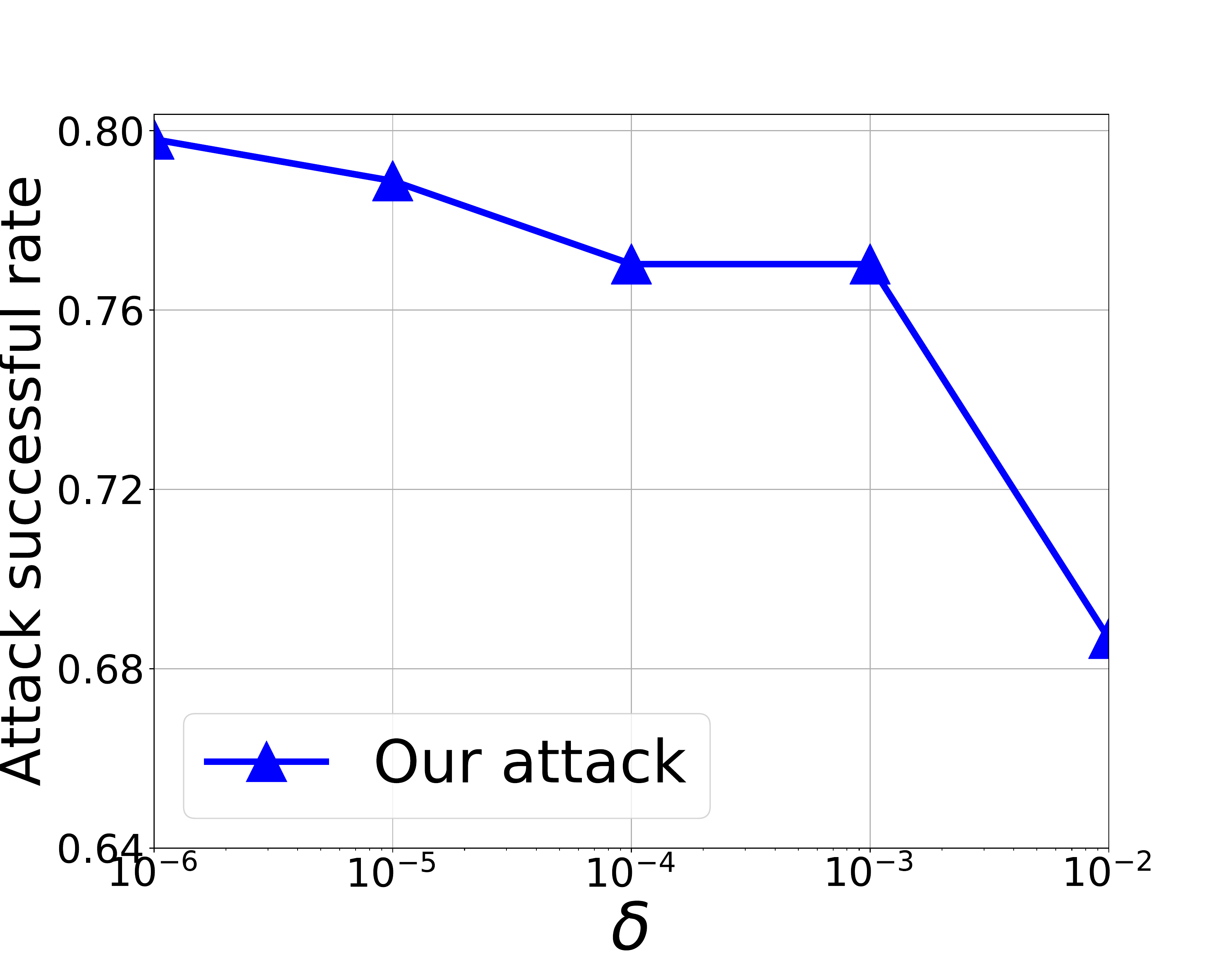}
\end{minipage}%
}%
\subfigure[SGC with Pubmed.]{
\begin{minipage}[t]{0.24\linewidth}
\centering
\includegraphics[width=\columnwidth]{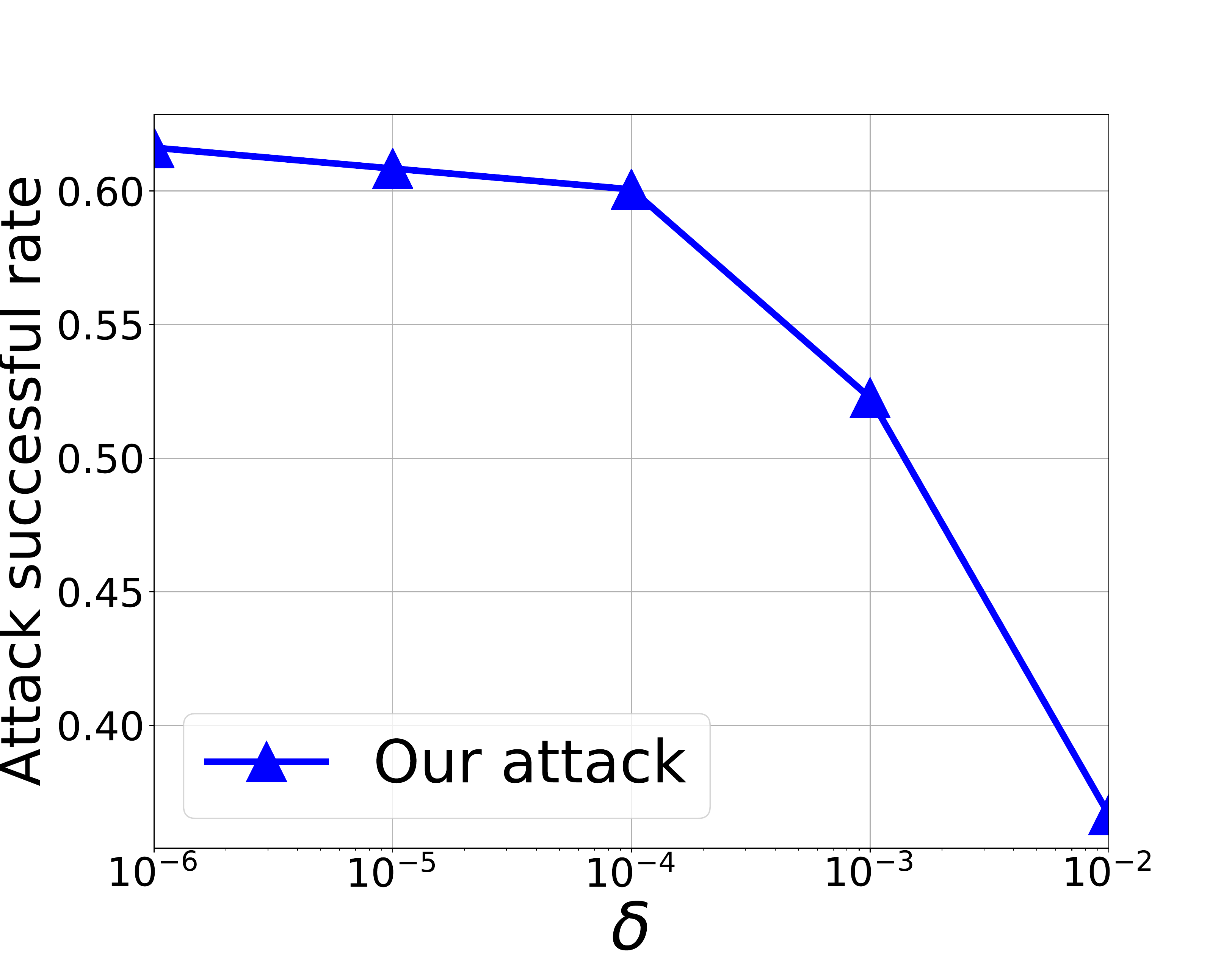}
\end{minipage}%
}%
\subfigure[GIN with CIFAR10.]{
\begin{minipage}[t]{0.24\linewidth}
\centering
\includegraphics[width=\columnwidth]{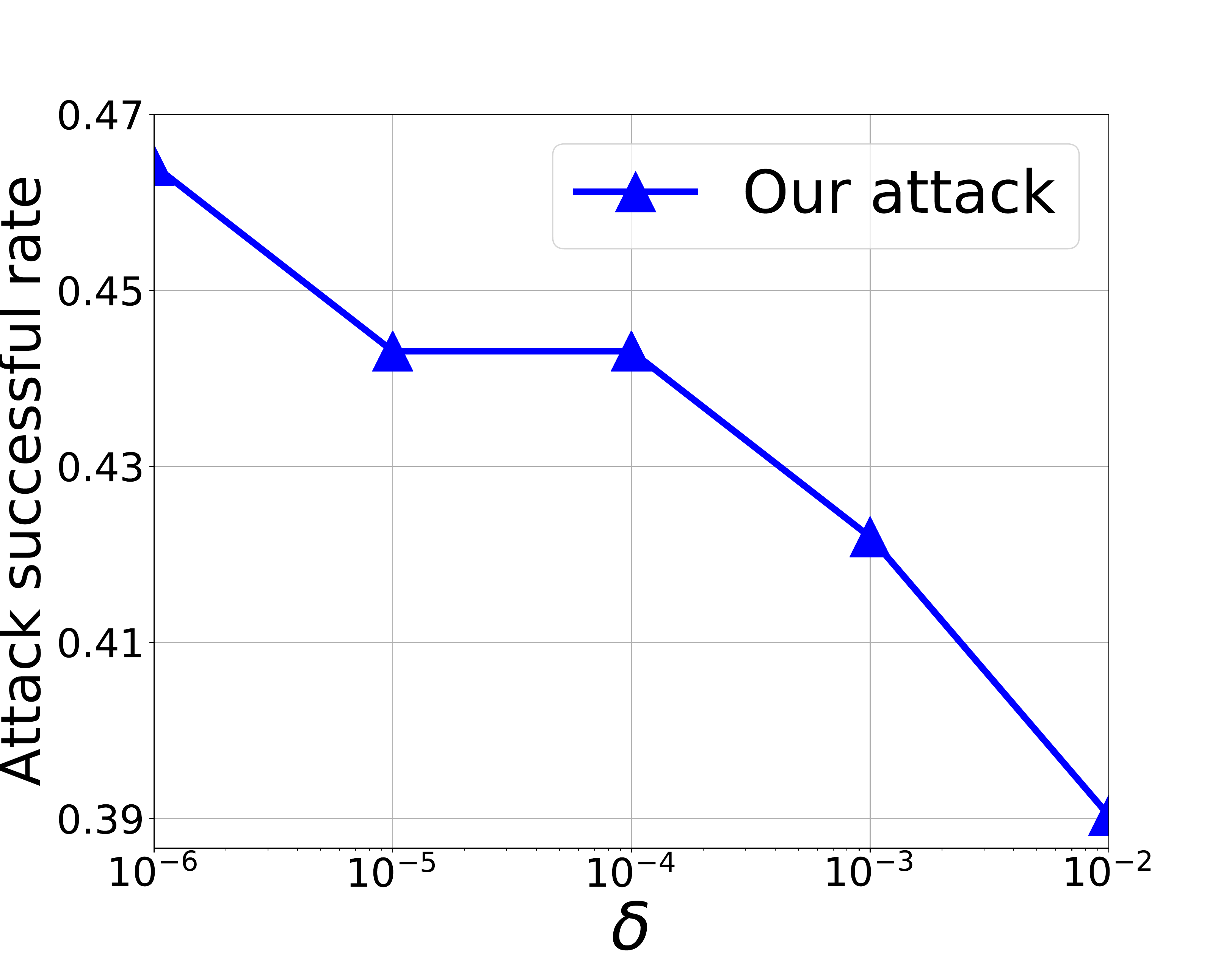}
\end{minipage}%
}%
\centering
\caption{Sensitivity of hyperparameter $\delta$.} 
\label{fig:delta}
\end{figure*}

\section{More Experimental Results}
In the following sections, we conduct experiments to evaluate the sensitivities of hyperparameters, i.e., projection scale $\alpha$ and update step $\delta$ of the prior vector. We set the default queries as $50$ for both node classification and graph classification, i.e., $T=50$. The default number of the perturbed edges is set to $2$ and $6$ for node classification and graph classification, respectively. Other parameter settings are consistently configured as the main experiments like learning rate $\eta$. $\alpha$ is ranged in $\{0,0.2,0.4,0.6,0.8\}$ and $\delta$ is ranged in $\{10^{-6},10^{-5},10^{-4},10^{-3},10^{-2}\}$. The results are shown in Figure \ref{fig:alpha} and Figure \ref{fig:delta}. Overall, the derivation of the attack successful rate is no more than $10\%$ under different hyperparameters, which demonstrates that our attack is stable and robust to sensitivity.

{\bf Sensitivity evaluation of hyperparameter $\alpha$.}
In Figure \ref{fig:alpha}, we evaluate the sensitivity performance of hyperparameter $\alpha$ over different GNNs (i.e., GCN, SGC and GIN) and different datasets (i.e., Cora, Citeseer, Pubmed, MNIST and CIFAR10). We can observe that the attack successful rate increases as $\alpha$ increases. This can be explained that the projected domain in PGD becomes larger when $\alpha$ goes larger. Thus, there is a high probability that the optimal perturbation is located in the larger projected domain. 

{\bf Sensitivity evaluation of hyperparameter $\delta$.}
In Figure \ref{fig:delta}, we evaluate the sensitivity performance of hyperparameter $\delta$ over different GNNs (i.e., GCN, SGC and GIN) and different dataset (i.e., Cora, Citeseer, Pubmed, MNIST and CIFAR10). We can observe that the attack successful rate decreases as $\delta$ increases. This can be explained that the estimated gradient $\bm{\hat{g}} = \frac{N}{\delta}L(\bm{\hat{s}}_v^t)\bm{u}^t$ becomes more inaccurate when $\delta$ goes larger. Consequently, it is impossible to derive the optimal perturbation with an inaccurate gradient. 

\end{document}